\ifdefined\pdfminorversion\pdfminorversion=7\fi
\RequirePackage{iftex}
\ifXeTeX
  \RequirePackage{fontspec}
  \let\TridentSetMonoFont\setmonofont
  \renewcommand{\setmonofont}[2][]{%
    \TridentSetMonoFont[#1,BoldFont=Inconsolatazi4-Bold.otf]{Inconsolatazi4-Regular.otf}}
\fi
\documentclass[sigconf,nonacm,balance=false]{acmart}
\ifXeTeX\let\setmonofont\TridentSetMonoFont\fi
\setcopyright{none}
\usepackage{multirow}
\usepackage{placeins}
\usepackage{makecell}
\usepackage{enumitem,tabularx}
\newcounter{pfcAlgorithm}
\newcommand{\pfcindent}{\hspace*{1.2em}}
\newcommand{\pmstd}[1]{{\scriptsize$\pm$#1}}
\AtBeginDocument{\hypersetup{pdfauthor={Weiyu Han, Haoyu Jiang, Jie Sheng, Ruobing Zuo, Huilin He, Ze Zhao, Xiao Tan, Nairong Zhang, Shuai Chen, Tianyi Zhang},pdfkeywords={online payments, payment risk modeling, transaction foundation models, self-supervised learning, contrastive learning, transfer learning}}}
\ifXeTeX
  \AddToHook{cmd/maketitle/after}{}
\fi
\title{TRIDENT-Seq: Profile-Matched Prefix--Future Contrast for Label-Efficient Payment Risk Modeling}
\author{Weiyu Han}
\affiliation{\institution{Ant International}\country{}}
\email{hanweiyu.hwy@ant-intl.com}
\author{Haoyu Jiang}
\affiliation{\institution{Ant International}\country{}}
\email{wuxian.jhy@ant-intl.com}
\author{Jie Sheng}
\affiliation{\institution{Ant International}\country{}}
\email{shengjie.sheng@ant-intl.com}
\author{Ruobing Zuo}
\affiliation{\institution{East China Normal University}\country{}}
\affiliation{\institution{Ant International}\country{}}
\email{rbzuo@stu.ecnu.edu.cn}
\author{Huilin He}
\affiliation{\institution{Tongji University}\country{}}
\affiliation{\institution{Ant International}\country{}}
\email{huilin3@tongji.edu.cn}
\author{Ze Zhao}
\affiliation{\institution{Shanghai Jiao Tong University}\country{}}
\affiliation{\institution{Ant International}\country{}}
\email{zhaoze@sjtu.edu.cn}
\author{Xiao Tan}
\affiliation{\institution{Ant International}\country{}}
\email{alex.tx@ant-intl.com}
\author{Nairong Zhang}
\affiliation{\institution{Ant International}\country{}}
\email{zhangnairong.znr@ant-intl.com}
\author{Shuai Chen}
\affiliation{\institution{Ant International}\country{}}
\email{shuai.cs@ant-intl.com}
\author{Tianyi Zhang}
\affiliation{\institution{Ant International}\country{}}
\email{zty113091@ant-intl.com}
\authorsaddresses{}
\renewcommand{\shortauthors}{Han et al.}
\makeatletter
\g@addto@macro\@authornotes{%
  \footnotetext[1]{Corresponding author: \href{mailto:alex.tx@ant-intl.com}{alex.tx@ant-intl.com}.}
  \footnotetext[2]{Work done during internships at Ant International.}}
\def\@mkauthors@iii{%
  \global\setbox\mktitle@bx=\vbox{\noindent\unvbox\mktitle@bx\par\bigskip
    \centering\normalfont\fontsize{12}{16}\selectfont
    Weiyu Han\textsuperscript{1}, Haoyu Jiang\textsuperscript{1}, Jie Sheng\textsuperscript{1}, Ruobing Zuo\textsuperscript{1,2,\textdagger}, Huilin He\textsuperscript{1,3,\textdagger}\par
    Ze Zhao\textsuperscript{1,4,\textdagger}, Xiao Tan\textsuperscript{1,*}, Nairong Zhang\textsuperscript{1}, Shuai Chen\textsuperscript{1}, Tianyi Zhang\textsuperscript{1}\par
    \medskip\fontsize{10}{13}\selectfont
    \textsuperscript{1}Ant International\quad
    \textsuperscript{2}East China Normal University\quad
    \textsuperscript{3}Tongji University\quad
    \textsuperscript{4}Shanghai Jiao Tong University\par\vskip14pt}}
\makeatother
\begin{abstract}
Ant International accumulates large volumes of transaction history across users and payment businesses. Confirmed fraud and chargeback outcomes, by contrast, are scarce, delayed, and specific to each business. A single encoder pretrained on unlabeled histories could serve all businesses, but only if pretraining captures the behavior that separates risk. Coarse activity differences can make contrastive pretraining on transaction logs too easy: transaction volume, amount, and timing can distinguish a history's continuation from arbitrary candidates without modeling how payment resources are used. We present TRIDENT-Seq, a transaction foundation model trained with Prefix--Future Contrast (PFC). PFC matches each prefix to a summary of its subsequent behavior among candidates with similar activity profiles; activity prediction and dispersion regularization act on the encoder output. Across three production payment businesses, the pretrained encoder improves average precision (AP) and recall at a 0.5\% intervention budget over architecture-matched scratch training in all six business--budget settings. With a fixed validation set, it exceeds full-label scratch training in mean AP and recall using 10\% of the training labels on Wallet and 25\% on Shopping and SME. Pretraining only on another business's histories still helps. Replacing profile-matched candidates with random ones cuts recall from 24.62\% to 5.64\%. In Ant International's deployment evaluation, recorded amount-based chargeback and overall loss rates are 37.7\% and 37.9\% lower than those of the previous XGBoost-based system under the same intervention budget.
\end{abstract}
\ccsdesc[300]{Applied computing~Secure online transactions}
\ccsdesc[500]{Computing methodologies~Unsupervised learning}
\ccsdesc[300]{Computing methodologies~Transfer learning}
\keywords{online payments, payment risk modeling, transaction foundation models, self-supervised learning, contrastive learning, transfer learning}
\begin{document}\flushbottom
\maketitle
\begin{figure*}[t]
  \centering
  \includegraphics[width=\textwidth]{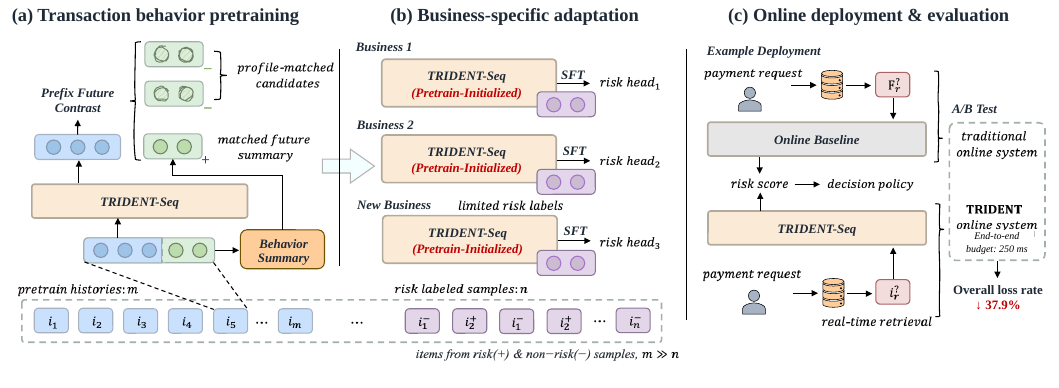}
  \caption{TRIDENT-Seq overview. (a) PFC learns behavioral representations without risk labels. (b) The shared foundation model adapts to each business through supervised fine-tuning (SFT) with limited labels. (c) At Ant International, online A/B evaluation across deployed payment businesses shows a 37.9\% reduction in the overall loss rate under the same intervention budget.}
  \Description{Three stages: behavior pretraining from prefixes and future summaries; reuse of the same pretrained encoder for business-specific supervised fine-tuning; and production risk evaluation.}
  \label{fig:overview}
\end{figure*}
\section{Introduction}
Online payments underpin commerce on the Web. Ant International supports payment services across digital wallets, shopping platforms, and merchant businesses. Each needs its own risk model. Transaction histories accumulate continuously in all of them, but confirmed risk outcomes are rare and arrive only after chargebacks or investigations~\cite{dalpozzolo2018}. A new business or market therefore starts with abundant behavior and few labels, and one model per business multiplies the cost of labeling and upkeep~\cite{pragma2026}. The natural remedy is to pretrain one encoder on unlabeled histories and adapt it to each business with whatever labels it has.

Such an encoder should meet two requirements. \textbf{Low-label adaptation:} a business may hold millions of histories but only a few hundred confirmed outcomes. The pretrained representation must therefore separate much of the risk before labels shape it. \textbf{Cross-business generalization:} payment processes, user populations, and risk criteria differ across businesses. What the encoder learns must remain useful outside the business that supplied its histories, including businesses where most transactions have little or no history.

Self-supervised learning on event sequences makes this feasible~\cite{padhi2021tabbert,babaev2022coles,oord2018cpc,wu2025byb}, and transaction foundation models now transfer representations across financial tasks and populations~\cite{skalski2023fpm,dou2025tgpt,li2025panther,pragma2026}. Less settled is what a pretraining task should demand of a payment history. Risk often shows in how a user handles resources over time: switching cards after a decline, or retrying at the same merchant within minutes. Volume alone says much less.

This distinction exposes a weakness of contrastive pretraining on transaction logs. Histories differ most visibly in activity level: how many transactions they contain, how large the amounts are, and how the events spread over time. Asked to identify the continuation of a history among arbitrary alternatives, a model can often succeed on these coarse quantities alone. It then has little pressure to learn the finer behavior that separates risky from legitimate use, an instance of the shortcut solutions known to affect contrastive learning~\cite{robinson2021shortcut}. Candidate construction has a large downstream impact in our data. On a wallet business, random contrastive candidates yield 5.64\% recall at a 0.5\% intervention budget after fine-tuning with 10\% of the labels. Candidates of comparable activity yield 24.62\% (Section~\ref{sec:ablation}).

We present TRIDENT-Seq, a transaction foundation model built on \emph{Prefix--Future Contrast} (PFC). PFC splits each training history into an older prefix and the behavior that follows. It summarizes the latter through amount, timing, and attribute-change statistics, and trains the encoder to identify the true summary among candidates with similar activity profiles. PFC selects candidates with comparable transaction counts, amounts, and aggregate event ages, then distinguishes their richer behavioral summaries. Two auxiliary terms act on the encoder output that adaptation reuses. Activity prediction directly supervises activity level, complementing the profile-matched contrast; a dispersion penalty discourages low average variation in the encoder output. Each business then fine-tunes a copy of the encoder with its own risk head (Figure~\ref{fig:overview}).

We evaluate TRIDENT-Seq on three payment businesses served by Ant International, with history coverage ranging from nearly complete to mostly empty. Pretraining improves AP and recall over architecture-matched scratch training in all six business--budget settings. On the wallet business, 10\% of the training labels, only 76 positive examples, outperform scratch training with all training labels under the same fixed validation set. On the other two, 25\% do so in mean AP and recall. Pretraining solely on another business's histories still helps. The system has served SME merchants, where most transactions have no history, since June 8, 2026. Across its deployed payment businesses, recorded chargeback and overall loss rates are 37.7\% and 37.9\% lower than those of the previous XGBoost-based system.

Our contributions are as follows:
\begin{itemize}[leftmargin=1.2em,itemsep=1pt,topsep=2pt]
\item \textbf{Method.} We propose profile-matched future contrast, which trains each history prefix to identify its own future summary among candidates with comparable activity. This makes candidate construction an explicit part of transaction pretraining.
\item \textbf{Evidence.} Across three production businesses, one pretrained encoder supports label-efficient adaptation and transfers across businesses, and ablations isolate the contribution of each component.
\item \textbf{Deployment.} TRIDENT-Seq runs in Ant International's payment risk system within a 120\,ms inference budget, with portfolio-level A/B results and a two-stage SME evaluation.
\end{itemize}

\section{Related Work}
\textbf{Transaction risk modeling.} Feature-based detectors aggregate historical activity into indicators for supervised classifiers such as gradient-boosted trees~\cite{bahnsen2016,dalpozzolo2018,chen2016xgboost}. Sequence models learn temporal dependencies directly from ordered transactions~\cite{jurgovsky2018}. Graph models exploit relations among accounts, devices, and merchants, including camouflaged and highly imbalanced fraud~\cite{liu2018gem,dou2020care,liu2021pcgnn}. These detectors typically learn from the risk labels of a single task. Each new business must first accumulate its own labels, and delayed feedback slows their arrival~\cite{dalpozzolo2018,carcillo2018scarff}. TRIDENT-Seq learns its representation from unlabeled histories and uses labels only for adaptation.

\textbf{Self-supervised pretraining on event sequences.} TabBERT reconstructs masked fields of tabular event sequences~\cite{padhi2021tabbert}, and self-supervised objectives have been combined with sampling strategies for card fraud detection~\cite{chen2024ccfd}. CoLES treats subsequences of the same history as positive pairs~\cite{babaev2022coles}, as TS2Vec does for time series~\cite{yue2022ts2vec}. Predictive objectives link the past to what follows. CPC contrasts future latent representations against in-batch negatives~\cite{oord2018cpc}, and BYB predicts a teacher's representation of a later behavior window~\cite{wu2025byb} in the spirit of BYOL~\cite{grill2020byol}. PFC is also predictive, but its target and its negatives differ. The target is a fixed statistical summary of several subsequent events, not a learned event representation. The negatives are restricted to comparable activity. Hard-negative sampling also selects informative negatives, but defines hardness by embedding similarity~\cite{robinson2021hard}. PFC defines comparability through observable activity profiles and thus targets the shortcut that random negatives leave open~\cite{robinson2021shortcut}. Finally, negatives keep contrastive representations spread over the hypersphere~\cite{wang2020align}, whereas non-contrastive methods enforce spread with explicit variance terms~\cite{bardes2022vicreg}. Our dispersion term follows the latter.

\textbf{Transaction foundation models.} The Foundation Purchasing Model pretrains generative autoregressive models on transaction sequences and transfers them to held-out fraud-detection populations~\cite{skalski2023fpm}. TransactionGPT models transaction attributes and trajectories with a 3D Transformer~\cite{dou2025tgpt}. PANTHER combines structured behavior tokens, sequence patterns, and user profiles~\cite{li2025panther}. PRAGMA extends masked modeling to banking events~\cite{pragma2026}, and FRiskGPT places user behaviors and risk outcomes in a shared token space~\cite{zhang2026friskgpt}. Payment companies report similar models in production~\cite{stripe2025,plaid2026a,plaid2026b}. Pretrained sequence encoders are likewise shared across domains in recommendation and advertising~\cite{hou2022unisrec,gong2023coldstart,fu2025ctr,liang2025exfm}, where transfer is usually evaluated between related catalogs, markets, or tasks~\cite{pan2010transfer}. Payment businesses also differ in label definitions and history coverage, which our controlled-source experiments vary explicitly. Most transaction foundation models predict the next event or reconstruct masked ones. TRIDENT-Seq instead contrasts multi-event summaries under activity matching, and we follow its reuse across businesses through to deployment.

\section{Problem Setup}
\label{sec:preliminary}
\textbf{Transaction histories.} For a transaction scored at time $t$, the history $H_{<t}=(e_1,\ldots,e_n)$ lists the user's preceding events from most recent to oldest. Each event $e_j=(a_j,\mathbf f_j,\delta_j)$ has an amount $a_j$, categorical attributes $\mathbf f_j$ describing payment, device, and network context, and an age $\delta_j$ in days relative to $t$. The length $n$ varies and can be zero. The label $y\in\{0,1\}$ of the scored transaction indicates recorded fraud or chargeback. Chargebacks can arrive months after the transaction, so labels always lag the traffic they describe. For pretraining, an observed history is split into an older prefix $H^p$ and the subsequent segment $H^f$, which we call its future. Both precede the scored transaction, so pretraining never sees events that occur after a decision.

\textbf{Pretraining and adaptation.} Let $\mathcal U$ be a pool of unlabeled histories and $\mathcal L_d=\{(H_{<t},y)\}$ the labeled transactions of business $d$. We learn encoder parameters $\theta_{\rm pre}$ on $\mathcal U$. Each business initializes its encoder $F_{\theta_d}$ from $\theta_{\rm pre}$ and trains it jointly with a risk head $g_{\phi_d}$:
\begin{equation}
\label{eq:risk-score}
p_d(H_{<t})=\sigma\!\left(g_{\phi_d}\!\left(F_{\theta_d}(H_{<t})\right)\right),
\end{equation}
where $\sigma$ is the sigmoid function. The decision system ranks transactions by $p_d$ and intervenes on a fixed fraction of traffic, the intervention budget. The budget is small, 0.5\% in Ant International's production system, because every intervention adds friction for legitimate users. We seek a $\theta_{\rm pre}$ that makes this ranking accurate with few labels, across businesses.

\section{Methodology}
\label{sec:method}
\subsection{Overview}
\label{sec:arch}
Figure~\ref{fig:overview} shows the three stages of TRIDENT-Seq. (a)~PFC pretrains a shared history encoder on unlabeled histories. (b)~Each business fine-tunes its own copy of the encoder, with a new risk head, on its own labels. (c)~Online, each transaction triggers history retrieval and one pass through the adapted encoder and risk head; the decision system acts on the resulting score. Future summaries, contrastive candidates, and auxiliary heads exist only in pretraining. They add nothing to serving cost.

\subsection{Event Encoding}
\textbf{Event tokenization.} Payment, device, and network attributes acquire new values continuously. We encode each categorical field with a Bloom-filter-style multi-probe hash~\cite{bloom1970,weinberger2009hashing,svenstrup2017hash} that maps a value to a fixed-width bit vector. A card or device always produces the same bit pattern. A new one typically produces a different pattern, and no vocabulary has to be maintained. Reuse and switching of payment resources, the cues behind evasion, are therefore visible directly in the token sequence. With four probes in separate 32-bit blocks, two distinct values receive identical vectors with probability $32^{-4}\approx10^{-6}$, so collisions rarely merge different cards or devices. Amounts are clipped and log-transformed, and ages are expanded with sine and cosine features~\cite{kazemi2019time2vec}. The token of event $e_j$ is
\begin{equation}
\mathbf x_j=P_e\!\left[B(\mathbf f_j)\,\|\,\psi(a_j)\right]+P_\delta\,\gamma(\delta_j),
\end{equation}
where $B$ concatenates the field bit vectors in a fixed order, $\psi$ transforms the amount, $\gamma$ expands the age, $\|$ denotes concatenation, and $P_e$ and $P_\delta$ are learned projections.

\begin{figure}[t]
  \centering
  \includegraphics[width=\columnwidth]{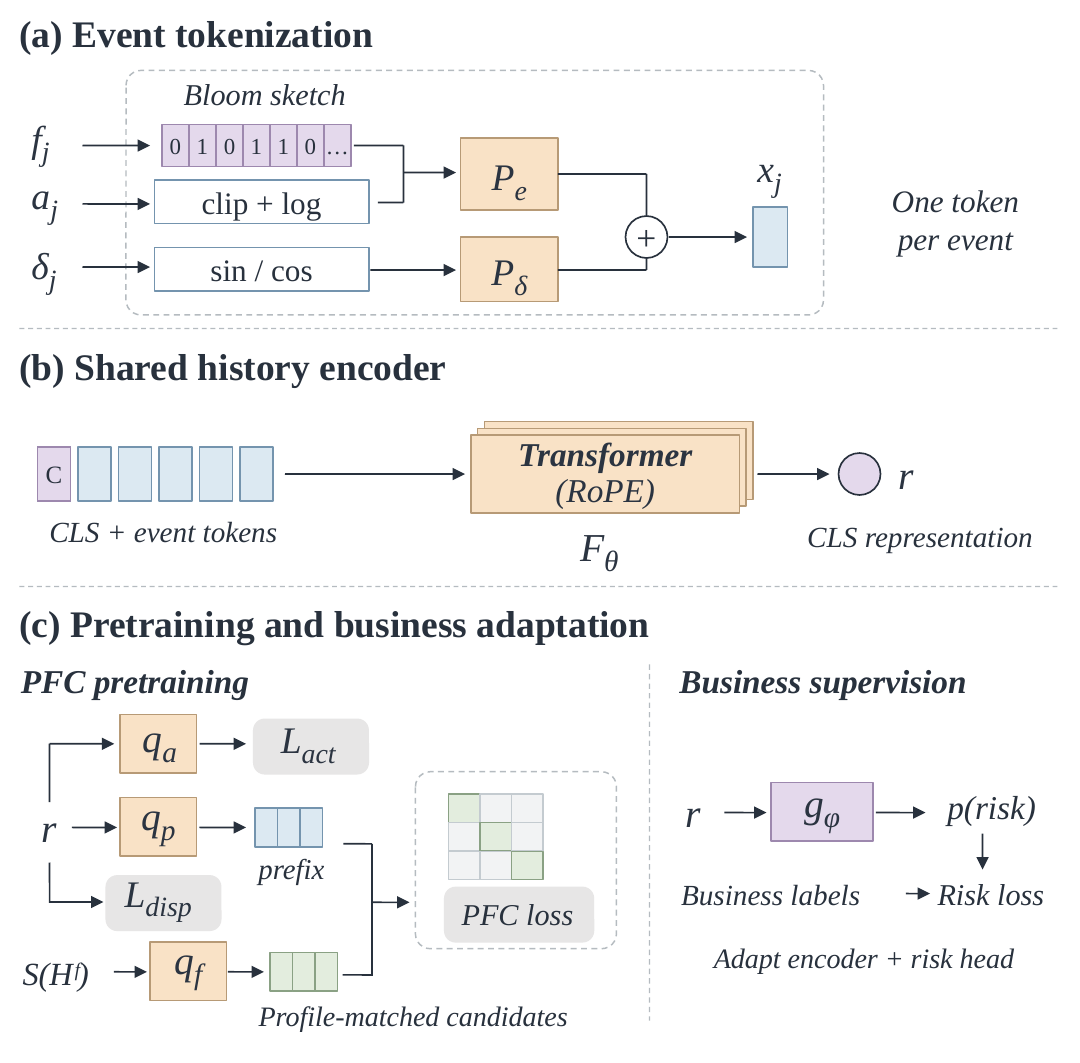}
  \caption{TRIDENT-Seq architecture. (a) Event tokenization. (b) Shared history encoder. (c) Pretraining with PFC, activity prediction, and dispersion regularization, and business-specific supervised adaptation.}
  \Description{Event attributes are encoded into tokens. A Transformer produces a history representation. Pretraining uses prefix and summary projection heads; downstream adaptation uses a business risk head.}
  \label{fig:arch}
\end{figure}

\textbf{History encoder.} A Transformer~\cite{vaswani2017} with rotary positional encoding reads a CLS token followed by the event tokens, with padding masked. Its CLS output $F_\theta(H)$ represents the history. Pretraining, adaptation, and serving share this encoding (Figure~\ref{fig:arch}); Appendix~\ref{app:impl} gives the configuration.

\subsection{Prefix--Future Contrast}
\label{sec:pfc}
\textbf{Prefixes and future summaries.} For a training history $H_i$ with at least two events, the older 60\% of events form the prefix $H_i^p$ and the rest form the future $H_i^f$. The encoder represents the prefix as $\mathbf r_i=F_\theta(H_i^p)$. A deterministic function $S$ maps $H_i^f$ to a 24-dimensional summary of its amount and event-age distributions, the regularity of intervals between adjacent events, and the occupancy and change of hashed attribute values (Appendix~\ref{app:targets}). Hash occupancy is a coarse proxy for the diversity of attribute values within the future. The switching rate measures, field by field, how often consecutive events change attribute values, as when a card or device is replaced after a rejection. Event-age statistics describe how activity is spread over time. We summarize several events rather than predict the next one: retry bursts, resource switching, and the slow build-up of routine activity all span multiple events. A prefix head $q_p$ and a future head $q_f$ map both sides into a shared space:
\begin{equation}
\mathbf z_i^p=\operatorname{norm}\!\left(q_p(\mathbf r_i)\right),\qquad
\mathbf z_i^f=\operatorname{norm}\!\left(q_f(S(H_i^f))\right),
\end{equation}
where $\operatorname{norm}$ is L2 normalization~\cite{chen2020simclr,wang2020align}.

\textbf{Profile-matched candidates.} Contrasting a prefix against arbitrary futures invites the activity shortcut: a future with three transactions is easily told apart from one with three hundred. We therefore select contrastive candidates by similarity in coarse activity. With $u_j=\log(1+\max(a_j,0))$, the activity profile of a future is
\begin{equation}
\label{eq:profile}
\pi(H^f)=
\begin{bmatrix}
\log\!\left(1+\sum_{j\in H^f}u_j\right)\\
\log(1+|H^f|)\\
\log\!\left(1+\sum_{j\in H^f}\delta_j\right)
\end{bmatrix}.
\end{equation}
In a minibatch of $B$ prefix--future pairs, prefix $i$ keeps its own future and every other future $j$ with $\|\pi(H_i^f)-\pi(H_j^f)\|_1\le\varepsilon$. If no other future qualifies, all futures in the batch are used (Algorithm~\ref{alg:pfc}). The logarithmic threshold $\varepsilon=1.5$ allows roughly $e^{1.5}\approx4.5$-fold variation in activity scale when the other two coordinates agree. Building the mask takes $B^2$ comparisons of three-dimensional profiles, negligible next to the encoder. With $\mathcal N(i)$ the resulting candidate set and $\tau$ a temperature, PFC minimizes
\begin{equation}
\mathcal L_{\mathrm{PFC}}=-\frac{1}{B}\sum_{i=1}^{B}
\log\frac{\exp(\langle\mathbf z_i^p,\mathbf z_i^f\rangle/\tau)}
{\sum_{j\in\mathcal N(i)}\exp(\langle\mathbf z_i^p,\mathbf z_j^f\rangle/\tau)}.
\end{equation}
Matching candidates on volume, amount, and event timing directs the contrast toward the finer structure that $S$ encodes.

\begin{table}[t]
\refstepcounter{pfcAlgorithm}\label{alg:pfc}
\caption*{\textbf{Algorithm \thepfcAlgorithm.} Profile-matched PFC for one minibatch}
\normalsize\setlength{\baselineskip}{10pt}\raggedright\hrule\smallskip
\textbf{Input:} $B$ histories with at least two events; prefix ratio $\rho$, threshold $\varepsilon$, temperature $\tau$.\par
\textbf{Output:} $\mathcal L_{\mathrm{PFC}}$ and prefix representations $R$.
\begin{enumerate}[leftmargin=1.65em,itemsep=0pt,parsep=0pt,topsep=2pt,label=\arabic*.]
\item \textbf{for} $i=1,\ldots,B$ \textbf{do}
\item \pfcindent $(H_i^p,H_i^f)\leftarrow\operatorname{SplitChronologically}(H_i,\rho)$;\ \ $\mathbf r_i\leftarrow F_\theta(H_i^p)$
\item \pfcindent $\mathbf z_i^p\leftarrow\operatorname{norm}(q_p(\mathbf r_i))$;\ \ $\mathbf z_i^f\leftarrow\operatorname{norm}(q_f(S(H_i^f)))$;\ \ $\pi_i\leftarrow\pi(H_i^f)$
\item $\ell_{ij}\leftarrow\langle\mathbf z_i^p,\mathbf z_j^f\rangle/\tau$ and $d_{ij}\leftarrow\|\pi_i-\pi_j\|_1$ for all $i,j$
\item $M\leftarrow I_B$;\ \ $M_{ij}\leftarrow1$ for all $j\ne i$ with $d_{ij}\leq\varepsilon$
\item $M_{i,:}\leftarrow\mathbf 1$ for every row $i$ with $\sum_j M_{ij}=1$ \hfill$\triangleright$ fallback
\item $\ell_{ij}\leftarrow-\infty$ wherever $M_{ij}=0$
\item \textbf{return} $\mathcal L_{\mathrm{PFC}}=-\frac1B\sum_i\log\operatorname{softmax}(\ell_{i,:})_i$,\ \ $R=[\mathbf r_1,\ldots,\mathbf r_B]$
\end{enumerate}\smallskip\hrule
\end{table}

\subsection{Auxiliary Objectives}
\label{sec:aux}
PFC trains the projected, normalized vectors $\mathbf z^p$, whereas adaptation reuses the encoder output $\mathbf r$. Two auxiliary terms act on $\mathbf r$ directly.

\textbf{Activity prediction.} Profile matching reduces coarse activity differences among candidates, while activity remains informative about risk. An activity head $q_a$ therefore regresses $\mathbf r_i$ onto a 25-dimensional target $A(H_i)$ computed from the full training history. The target collects transaction counts, amounts, event ages, and hash-field statistics over the 75, 150, and 300 most recent events, plus differences that contrast the most recent window with the longer history (Appendix~\ref{app:targets}). The loss
\begin{equation}
\label{eq:activity}
\mathcal L_{\mathrm{act}}=\frac1B\sum_{i=1}^{B}\operatorname{SmoothL1}\!\left(q_a(\mathbf r_i),A(H_i)\right)
\end{equation}
averages over target coordinates.

\textbf{Dispersion regularization.} Because the contrastive loss is computed after projection and normalization, it does not directly constrain the spread of $\mathbf r$ itself. We penalize a low average within-batch standard deviation of the prefix representations:
\begin{equation}
\label{eq:disp}
\mathcal L_{\mathrm{disp}}=\max\Bigl(0,\ \sigma_0-\frac1D\sum_{k=1}^{D}\operatorname{Std}_{i}\bigl(r_{ik}\bigr)\Bigr),
\end{equation}
where $\operatorname{Std}_i$ is the standard deviation of dimension $k$ over the $B$ prefixes in the minibatch and $D$ is the representation width. The penalty vanishes once the average spread reaches $\sigma_0$. It resembles the variance term of VICReg~\cite{bardes2022vicreg} but applies to the mean standard deviation rather than to each dimension, before projection.

\textbf{Joint objective.} The pretraining loss is
\begin{equation}
\label{eq:pretrain}
\mathcal L=\mathcal L_{\mathrm{PFC}}+\lambda_{\mathrm{act}}\mathcal L_{\mathrm{act}}+\lambda_{\mathrm{disp}}\mathcal L_{\mathrm{disp}},
\end{equation}
with $\lambda_{\mathrm{act}}=0.30$, $\lambda_{\mathrm{disp}}=0.05$, and $\sigma_0=0.02$. The encoder and the prefix, future, and activity heads are trained jointly.

\subsection{Business-Specific Adaptation}
\label{sec:adapt}
Each business $d$ copies the complete pretrained encoder, including event projections: $\theta_d\leftarrow\theta_{\rm pre}$. A new risk head $g_{\phi_d}$ reads its CLS output (Equation~\ref{eq:risk-score}), and the pretraining heads are discarded. For a labeled subset $\mathcal L_d$ with $N_d$ examples, $N_{d,c}$ of them in class $c$, the encoder and risk head minimize class-balanced cross-entropy:
\begin{equation}
\begin{split}
\mathcal L_{\rm risk}^{(d)}=-\frac{1}{N_d}\sum_{(H,y)\in\mathcal L_d}
\bigl[&\omega_{d,1}y\log p_d(H)\\
&+\omega_{d,0}(1-y)\log(1-p_d(H))\bigr],
\end{split}
\end{equation}
with $\omega_{d,c}=N_d/(2N_{d,c})$, so both classes carry equal total weight. We fine-tune the whole encoder rather than a head on frozen features. Each business defines risk differently, and its labels should be able to reweight what the shared representation emphasizes.

\section{Experiments}
We address four research questions. \textbf{RQ1:} Does pretraining improve risk ranking across businesses and label budgets? \textbf{RQ2:} Does it transfer when the target business is excluded from pretraining? \textbf{RQ3:} Which components of PFC matter? \textbf{RQ4:} Do the gains carry over to production?

\subsection{Datasets}
We use data from three payment businesses served by Ant International: a wallet business (Wallet), a shopping-platform business (Shopping), and an SME-merchant business (SME). They differ mainly in history coverage (Table~\ref{tab:data}). Nearly every Wallet transaction has a history, whereas 30.63\% of Shopping and 76.91\% of SME test traffic have none. Wallet is also the most label-scarce: its training split contains 732 positive examples, against 183,398 on Shopping and 46,740 on SME. Each dataset is split chronologically into training, validation, and test periods~\cite{dalpozzolo2018,carcillo2018scarff}, and transactions with empty histories remain in the evaluation. Shared pretraining uses a pool of 24.5 million unlabeled histories, including Shopping histories. Their anchors end on March 1, 2026, before every downstream test period.

In Table~\ref{tab:data}, scored and history events are counted over the sampled training, validation, and test sets. A history event is counted each time it appears in an input, so events shared by overlapping histories count repeatedly. Mean length includes empty histories. History sparsity is the population-weighted share of test traffic with an empty history. Appendix~\ref{app:data} lists split sizes and positive counts.

\begin{table}[t]
\caption{Dataset statistics.}
\label{tab:data}
\small
\setlength{\tabcolsep}{3pt}
\begin{tabular*}{\columnwidth}{@{\extracolsep{\fill}}lrrrrr@{}}
\toprule
Data & \makecell[r]{Scored\\events} & \makecell[r]{History\\events} & \makecell[r]{Mean\\length} & \makecell[r]{History\\sparsity} & \makecell[r]{Train\\positives}\\
\midrule
Wallet & 127,375 & 16.03M & 125.8 & 0.50\% & 732\\
Shopping & 501,898 & 14.78M & 29.5 & 30.63\% & 183,398\\
SME & 371,600 & 2.91M & 7.8 & 76.91\% & 46,740\\
\bottomrule
\end{tabular*}
\end{table}

\subsection{Experimental Setup}
\label{sec:setup}
\textbf{Metrics.} We report average precision (AP)~\cite{saito2015pr,davis2006pr} and risk recall at a 0.5\% intervention budget (R@0.5\%), the share of risk captured when the highest-scoring 0.5\% of traffic is selected. Validation and test metrics use inverse sampling-probability weights $w_i$ that account for cohort subsampling and negative sampling~\cite{horvitz1952,keilwagen2014}. For budget $b$, let $\alpha_i(b)$ be 1 above the score threshold, 0 below it, and a common fraction for scores tied at the threshold. Then
\begin{equation}
\mathrm{R}@b=\frac{\sum_i w_i y_i\alpha_i(b)}{\sum_i w_i y_i},
\qquad\sum_i w_i\alpha_i(b)=b\sum_i w_i.
\end{equation}
AP uses the same weights. Both metrics are reported as percentages. R@0.5\% is our primary metric. It mirrors how the decision system acts, intervening on a fixed share of traffic, and the 0.5\% budget matches production (Section~\ref{sec:online}). AP complements it by summarizing precision over the whole ranking.

\textbf{Label budgets.} Fine-tuning uses 1\%, 5\%, 10\%, 25\%, or 100\% of the training labels, with the same subsets for all methods. The validation set, which selects learning rates and checkpoints, is the same at every budget; on Wallet it holds 807 positives. The main comparison uses 10\% and 100\%, and Shopping and SME are evaluated at all five budgets. On Wallet, the 1\% and 5\% subsets would contain only 7 and 34 positive examples, so we report 10\% (76 positives) and 100\%.

\textbf{Baselines.} All offline methods take historical transaction events as input. \textbf{Scratch} uses the TRIDENT-Seq encoder and risk head with random initialization. \textbf{CoLES}~\cite{babaev2022coles}, \textbf{BYB}~\cite{wu2025byb}, and \textbf{CPC}~\cite{oord2018cpc} are adapted to share our encoder, pretraining histories, and 24,000-update budget. They contrast history crops, predict a teacher's representation of later behavior, and contrast upcoming event representations, respectively. \textbf{TransactionGPT}~\cite{dou2025tgpt} and \textbf{PANTHER}~\cite{li2025panther} keep their own event encodings and architectures and are pretrained on the same pool. Appendix~\ref{app:systems} details the adaptations and also reports XGBoost on production risk indicators, whose inputs differ from those of the sequence models. Ant International's production baselines, XGB-AutoRefit and a SAINT-style model, appear in Section~\ref{sec:online}.

\textbf{Implementation.} The encoder is an 8-layer Transformer with hidden size 1,024 and 8 attention heads (92.92M parameters) over at most 300 events. Pretraining uses AdamW at learning rate $1.5\times10^{-4}$, $\tau=0.2$, and $\varepsilon=1.5$ for 24,000 updates at global batch size 128. One pretrained checkpoint serves all businesses and budgets. Each fine-tuning run tries three encoder learning rates for 1,024 updates each, then trains for 2,048 updates at the rate with the best weighted validation AP. The same criterion selects the checkpoint. We report means and sample standard deviations over three fine-tuning runs. Appendix~\ref{app:impl} gives the full configuration, and Appendix~\ref{app:extended} examines a longer adaptation schedule.

\begin{table*}[t]
\caption{Performance comparison (\%). Best results are bold; second-best results are underlined.}
\label{tab:main}
\centering\small
\setlength{\tabcolsep}{3pt}
\begin{tabular*}{\textwidth}{@{\extracolsep{\fill}}lrrrrrr@{}}
\toprule
& \multicolumn{2}{c}{Wallet} & \multicolumn{2}{c}{Shopping} & \multicolumn{2}{c}{SME}\\
\cmidrule(lr){2-3}\cmidrule(lr){4-5}\cmidrule(lr){6-7}
Model & AP $\uparrow$ & R@0.5\% $\uparrow$ & AP $\uparrow$ & R@0.5\% $\uparrow$ & AP $\uparrow$ & R@0.5\% $\uparrow$\\
\midrule
\multicolumn{7}{@{}l}{\emph{10\% training labels}}\\
Scratch & 0.0342\pmstd{0.0082} & 3.31\pmstd{0.77} & 0.3067\pmstd{0.0375} & 4.79\pmstd{0.51} & 0.3476\pmstd{0.0798} & 4.67\pmstd{0.51}\\
CoLES (adapted) & 0.0274\pmstd{0.0124} & 2.92\pmstd{0.55} & 0.2879\pmstd{0.0182} & 5.18\pmstd{0.84} & 0.3514\pmstd{0.0691} & 5.04\pmstd{0.52}\\
BYB (adapted) & 0.0475\pmstd{0.0160} & 2.86\pmstd{0.56} & 0.2447\pmstd{0.0574} & 3.56\pmstd{1.80} & 0.3624\pmstd{0.0488} & 5.91\pmstd{2.17}\\
CPC (adapted) & 0.0279\pmstd{0.0061} & 3.34\pmstd{2.14} & 0.3569\pmstd{0.0094} & 5.79\pmstd{0.70} & 0.3874\pmstd{0.0283} & 5.60\pmstd{0.68}\\
TransactionGPT (adapted) & \underline{0.3209}\pmstd{0.1502} & \underline{20.69}\pmstd{8.57} & \underline{0.4316}\pmstd{0.0237} & \underline{6.16}\pmstd{0.88} & \textbf{0.5446}\pmstd{0.1264} & \underline{8.30}\pmstd{1.72}\\
PANTHER (adapted) & 0.2067\pmstd{0.1055} & 16.01\pmstd{9.69} & 0.2561\pmstd{0.0652} & 3.93\pmstd{1.13} & 0.3662\pmstd{0.0913} & 5.50\pmstd{1.01}\\
\textbf{TRIDENT-Seq} & \textbf{0.8214}\pmstd{0.3337} & \textbf{27.75}\pmstd{5.44} & \textbf{0.4331}\pmstd{0.0735} & \textbf{6.61}\pmstd{1.16} & \underline{0.4886}\pmstd{0.0672} & \textbf{9.14}\pmstd{1.85}\\
\midrule
\multicolumn{7}{@{}l}{\emph{100\% training labels}}\\
Scratch & 0.1148\pmstd{0.0486} & 10.41\pmstd{5.19} & 0.4205\pmstd{0.0660} & 7.81\pmstd{1.76} & 0.6370\pmstd{0.0805} & 8.11\pmstd{1.25}\\
CoLES (adapted) & 0.1150\pmstd{0.1515} & 5.87\pmstd{3.23} & 0.4919\pmstd{0.0605} & 8.25\pmstd{0.77} & 0.3865\pmstd{0.0268} & 6.10\pmstd{0.84}\\
BYB (adapted) & 0.0525\pmstd{0.0199} & 6.76\pmstd{2.12} & 0.5087\pmstd{0.0211} & 9.24\pmstd{0.34} & \underline{0.7682}\pmstd{0.2357} & 9.80\pmstd{1.47}\\
CPC (adapted) & 0.3128\pmstd{0.0968} & 12.74\pmstd{2.74} & 0.5057\pmstd{0.0497} & 7.48\pmstd{3.47} & 0.7678\pmstd{0.1896} & 10.89\pmstd{0.10}\\
TransactionGPT (adapted) & \underline{0.4447}\pmstd{0.1306} & \underline{21.33}\pmstd{5.45} & \textbf{0.7500}\pmstd{0.3085} & \textbf{11.39}\pmstd{1.51} & 0.7556\pmstd{0.0761} & \underline{14.20}\pmstd{0.37}\\
PANTHER (adapted) & 0.3484\pmstd{0.1383} & 16.09\pmstd{0.62} & 0.3074\pmstd{0.0281} & 5.18\pmstd{0.17} & 0.4573\pmstd{0.0386} & 6.61\pmstd{1.00}\\
\textbf{TRIDENT-Seq} & \textbf{0.8355}\pmstd{0.4142} & \textbf{28.33}\pmstd{3.25} & \underline{0.5678}\pmstd{0.0487} & \underline{9.61}\pmstd{0.73} & \textbf{0.8685}\pmstd{0.1553} & \textbf{14.34}\pmstd{0.42}\\
\bottomrule
\end{tabular*}
\end{table*}

\subsection{Overall Performance (RQ1)}
\label{sec:overall}
\textbf{Pretraining improves on scratch training in every setting.} TRIDENT-Seq raises both AP and R@0.5\% over architecture-matched scratch training in all six business--budget settings (Table~\ref{tab:main}). No other objective that shares our encoder and pretraining data does so. With 10\% of the Wallet labels, CoLES and CPC fall below scratch in AP, and BYB falls below it in recall. CPC is the closest to PFC: it also predicts what follows a prefix, but contrasts against in-batch futures of other users without activity matching. With full Wallet labels it beats scratch (12.74\% vs.\ 10.41\% recall) yet stays far from TRIDENT-Seq (28.33\%), in line with the drop that random candidates cause in our ablation (Section~\ref{sec:ablation}). Among the separately designed systems, PANTHER is competitive on Wallet but trails scratch on Shopping at both budgets and on SME with full labels. One difference: PANTHER encodes behaviors and fields through fixed vocabularies of 60,003 and 7,912 tokens, whereas hashing encodes unseen attribute values without a fixed vocabulary. For reference, XGBoost trained on Ant International's production risk indicators that preceded TRIDENT-Seq online trails scratch training in all six settings (Appendix~\ref{app:systems}).

\textbf{Wallet benefits most.} Histories cover 99.50\% of Wallet test traffic and average 125.8 events. With 10\% labels, TRIDENT-Seq reaches 0.8214\% AP and 27.75\% recall, 24.0 and 8.4 times the scratch values. Both exceed scratch trained on all labels (0.1148\% and 10.41\%). TRIDENT-Seq also leads the strongest baseline, TransactionGPT, at both budgets. Its 10\%-label result already retains 98\% of its full-label AP and recall, whereas scratch training with 10\% labels retains only 32\% of its full-label recall. Wallet AP varies widely across fine-tuning runs, with standard deviations of 41\% and 50\% of the mean. The training split holds only 732 positives, and a few weighted positives decide the top of the ranking. Recall is steadier, at 20\% and 11\%.

\textbf{Shopping and SME are more competitive.} TransactionGPT, which predicts the fields of the next event, leads both metrics on full-label Shopping and AP on SME with 10\% labels. TRIDENT-Seq leads recall with 10\% labels on both businesses, and both metrics on SME with full labels. Two of these comparisons fall within run-to-run variation: Shopping AP with 10\% labels (0.4331\% vs.\ 0.4316\%) and SME recall with full labels (14.34\% vs.\ 14.20\%). SME deserves a closer look. Of its test traffic, 76.91\% has no history, yet pretraining nearly doubles recall with 10\% labels, from 4.67\% to 9.14\%. The reason is concentration: the history-covered 23.09\% of traffic carries 85.51\% of the weighted risk.

\textbf{The advantage persists under longer adaptation.} Each fine-tuning run above uses 5,120 updates, learning-rate search included. On full-label Wallet we extend this to 14,336 updates for both models. TRIDENT-Seq then reaches 2.53 times the recall and 1.79 times the AP of scratch training (Appendix~\ref{app:extended}), close to the recall ratio of 2.72 under the standard schedule.

\subsection{Label Efficiency (RQ1)}
\label{sec:label}
\begin{figure}[t]
\centering
\includegraphics[width=\columnwidth]{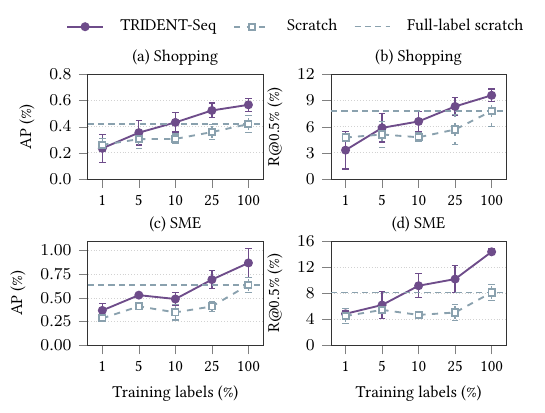}
\caption{Label efficiency on Shopping and SME.}
\Description{Four plots show AP and recall at a 0.5 percent intervention budget for TRIDENT-Seq and scratch training across five training-label budgets on Shopping and SME, with a dashed line for full-label scratch.}
\label{fig:label}
\end{figure}

Figure~\ref{fig:label} sweeps five label budgets on Shopping and SME. Points and error bars show three-run means and standard deviations; dashed lines mark scratch training with all labels. With 25\% of the training labels, TRIDENT-Seq exceeds full-label scratch in mean AP and recall on both businesses. On Shopping it reaches 0.5246\% AP and 8.33\% recall, against 0.4205\% and 7.81\%. On SME it reaches 0.6928\% and 10.16\%, against 0.6370\% and 8.11\%.

Pretraining also flattens the label curve. With 25\% of the labels, TRIDENT-Seq keeps 87\% of its full-label recall on Shopping and 71\% on SME; scratch training keeps 73\% and 62\%. TRIDENT-Seq improves both metrics over same-budget scratch at every budget on SME and from 5\% upward on Shopping. At 1\% of the Shopping labels, it trails scratch. The relative gain in both metrics peaks at 25\% on both businesses: 46\% in AP and 47\% in recall on Shopping, 69\% and 102\% on SME. With full labels it narrows as scratch training catches up. Appendix~\ref{app:budgets} lists all values.

\subsection{Cross-Business Transfer (RQ2)}
\label{sec:sources}
The main pretraining pool includes Shopping histories. To test transfer without target-business data~\cite{pan2010transfer}, we pretrain separate checkpoints on Shopping-only histories, SME-only histories, or an equal mixture. Each pool holds 1,288,542 distinct histories with matched daily counts. Each checkpoint receives 24,000 updates, and adaptation follows the main protocol at 10\% and 25\% labels.

Pretraining only on the other business improves AP and recall over no pretraining in all four target--budget settings (Table~\ref{tab:transfer}). The two directions differ. For Shopping, SME histories match Shopping's own in recall at 10\% and beat them in both metrics at 25\%. For SME, Shopping histories help more than SME's own with 10\% labels (0.5791\% vs.\ 0.4857\% AP; 8.67\% vs.\ 6.13\% recall) but less with 25\%. SME histories are short, averaging 7.8 events against 29.5 on Shopping, so richer histories from another business can partly substitute for in-domain ones. Mixing the two sources beats target-only pretraining in both metrics on both businesses at 25\% labels; at 10\%, it raises recall but lowers AP. Bold in Table~\ref{tab:transfer} marks the best mean per target and budget, and Appendix~\ref{app:sourcecontrols} adds standard deviations and AUROC.

\begin{table}[t]
\caption{Effect of the pretraining source (\%).}
\label{tab:transfer}
\centering\small
\setlength{\tabcolsep}{2pt}
\begin{tabular*}{\columnwidth}{@{\extracolsep{\fill}}lrrrr@{}}
\toprule
&\multicolumn{2}{c}{10\% labels}&\multicolumn{2}{c}{25\% labels}\\
\cmidrule(lr){2-3}\cmidrule(lr){4-5}
Pretraining source & AP & R@0.5\% & AP & R@0.5\%\\
\midrule
\multicolumn{5}{@{}l}{\textit{Target: Shopping}}\\
\hspace{1em}No pretraining & 0.3067 & 4.79 & 0.3591 & 5.68\\
\hspace{1em}Shopping only & \textbf{0.3994} & 5.97 & 0.4361 & 7.85\\
\hspace{1em}SME only & 0.3668 & 5.96 & \textbf{0.4663} & \textbf{8.70}\\
\hspace{1em}Shopping + SME & 0.3331 & \textbf{6.46} & 0.4511 & 7.98\\
\midrule
\multicolumn{5}{@{}l}{\textit{Target: SME}}\\
\hspace{1em}No pretraining & 0.3476 & 4.67 & 0.4103 & 5.04\\
\hspace{1em}SME only & 0.4857 & 6.13 & 0.5095 & 7.86\\
\hspace{1em}Shopping only & \textbf{0.5791} & \textbf{8.67} & 0.4319 & 6.94\\
\hspace{1em}Shopping + SME & 0.4315 & 7.15 & \textbf{0.6379} & \textbf{9.05}\\
\bottomrule
\end{tabular*}
\end{table}

\subsection{Ablation Study (RQ3)}
\label{sec:ablation}
We ablate PFC on Wallet, whose histories are the most complete and where pretraining matters most (Table~\ref{tab:ablation}). Each variant modifies Full PFC in one respect. Two change how the contrast is built. \emph{Random negatives} keeps each row's candidate count but draws candidates without profile matching. \emph{Shuffled pairs} keeps the prefixes and summaries but pairs each prefix with another history's future. Three change the auxiliary terms: \emph{w/o activity} and \emph{w/o dispersion} remove one term, and \emph{PFC only} removes both. All variants share the encoder, data, and 24,000 pretraining updates, and each is pretrained and fine-tuned once with the learning rate selected on full labels.

\textbf{Profile matching matters most.} Random negatives give the lowest recall at the 0.5\% and 1\% budgets: 5.64\% and 7.38\% at 0.5\%. They fall below even shuffled pairs, whose targets carry no information about the prefix. At these budgets, random negatives with full labels stay below every other variant with only 10\% of the labels. We read this as evidence for the activity shortcut of Section~\ref{sec:pfc}: when candidates can be separated by volume, pretraining appears to favor cues that transfer poorly to risk ranking. The ablation measures downstream recall; probing which cues the encoder relies on would test this reading directly.

\textbf{Correspondence matters.} Shuffled pairs reduce R@0.5\% to 14.62\% and 16.22\%, about the level of PFC only, and give the lowest recall of all variants at the 0.1\% budget. Comparable candidates help only when the prefix must find its own future among them. Without the true target, the contrast cannot teach how a particular history continues.

\textbf{Combining the auxiliary terms gives the highest recall.} Adding either term alone to PFC only raises R@0.5\% by at most 1.8 points with 10\% labels: from 14.37\% to 15.10\% with dispersion and 16.14\% with activity. Adding both raises it to 24.62\%. With full labels the pattern repeats: 16.38\%, 17.21\%, and 28.44\%, from 16.07\%. Activity prediction supplies explicit statistical supervision, while dispersion regularization penalizes low average variation in the encoder outputs. Their combination achieves the largest recall gain in Table~\ref{tab:ablation}.

The two removals also differ across budgets. Removing dispersion causes its largest drop at the 0.5\% budget, 8--11 points, against at most 4 points at 0.1\% and 1\%. Removing activity prediction lowers recall by 10--12 points at both 0.5\% and 1\%.

\textbf{The components mainly shape the top of the ranking.} At the 0.1\% budget, PFC only and the single-term variants stay within 2 points of Full PFC. AUROC, which summarizes the whole ranking, differs by less than 2.5 points across these variants (Appendix~\ref{app:ablation}). The gains concentrate at the 0.5\% and 1\% budgets, around the production operating point.

\begin{table}[t]
\caption{Wallet ablation: recall (\%) at three intervention budgets.}
\label{tab:ablation}
\centering\small
\setlength{\tabcolsep}{1.5pt}
\begin{tabular*}{\columnwidth}{@{\extracolsep{\fill}}lrrrrrr@{}}
\toprule
&\multicolumn{3}{c}{10\% labels}&\multicolumn{3}{c}{100\% labels}\\
\cmidrule(lr){2-4}\cmidrule(lr){5-7}
Variant / budget & 0.1\% & 0.5\% & 1.0\% & 0.1\% & 0.5\% & 1.0\%\\
\midrule
Full PFC & \textbf{8.54} & \textbf{24.62} & \textbf{32.70} & \textbf{11.26} & \textbf{28.44} & \textbf{36.29}\\
\midrule
Random negatives & 3.77 & 5.64 & 8.98 & 3.74 & 7.38 & 13.10\\
Shuffled pairs & 1.32 & 14.62 & 24.79 & 2.45 & 16.22 & 20.04\\
PFC only & 7.02 & 14.37 & 20.65 & 9.33 & 16.07 & 21.28\\
w/o activity & 7.85 & 15.10 & 22.50 & 10.31 & 16.38 & 24.07\\
w/o dispersion & 7.34 & 16.14 & 30.21 & 9.76 & 17.21 & 32.65\\
\bottomrule
\end{tabular*}
\end{table}

\subsection{Case Study}
\label{sec:case}
\begin{figure}[t]
  \centering
  \includegraphics[width=\columnwidth]{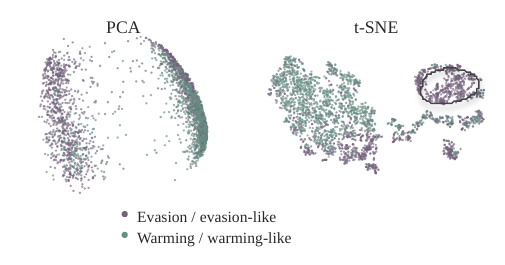}
  \caption{Pretrained representations of annotated histories.}
  \Description{PCA and t-SNE projections of pretrained representations, with two behavior groups and an outlined t-SNE region used for case inspection.}
  \label{fig:proj}
\end{figure}\begin{figure}[t]
  \centering
  \includegraphics[width=\columnwidth]{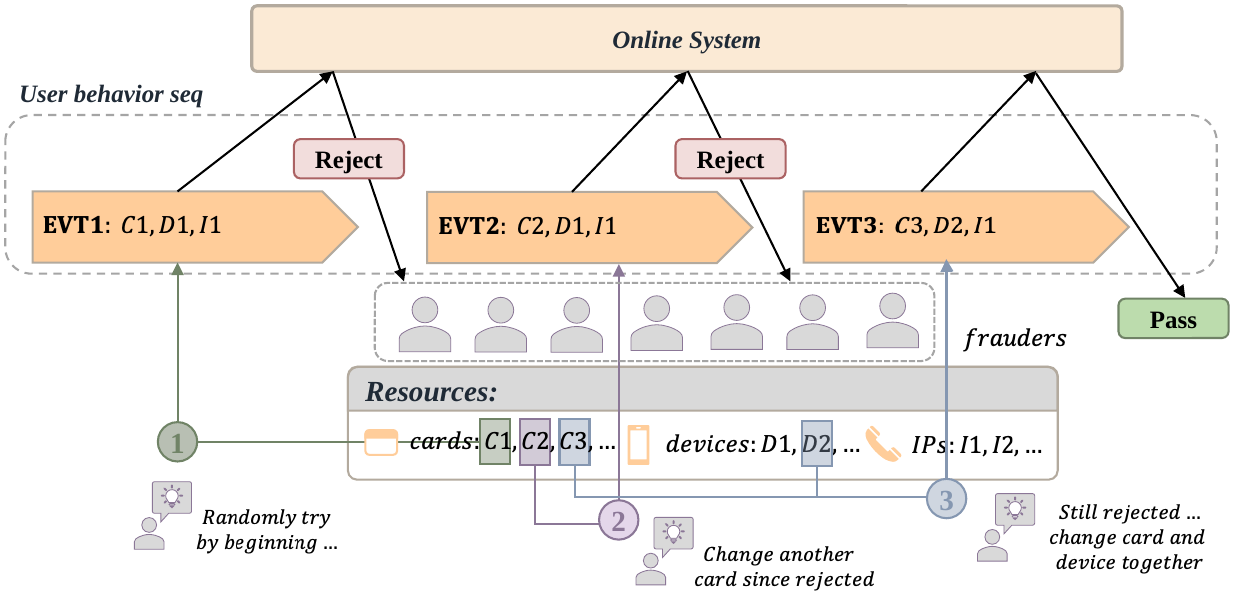}
  \caption{Card and device switching after rejection.}
  \Description{Three payment attempts show changes in anonymous card and device identities following rejections while retaining the IP identity, ending in a successful attempt.}
  \label{fig:retry}
\end{figure}
\textbf{Evasion and account warming.} These two risk patterns unfold on different time scales. Evasion changes payment resources or attempt timing to get past risk controls; it shows up as rapid adjustments across consecutive attempts. Account warming builds routine-looking activity before later misuse; it shows up as slow accumulation. We inspect the pretrained representations of 2,880 annotated histories. They include evasion and account-warming cases, together with legitimate histories that look similar. Figure~\ref{fig:proj} colors evasion with evasion-like legitimate histories, and account warming with warming-like ones. In PCA, the evasion group concentrates on the left branch and the warming group on the right. The two overlap. In t-SNE~\cite{maaten2008tsne}, the outlined region of 374 histories is dominated by evasion-type histories.

\textbf{Rapid retries at the same merchant characterize this region.} Among its 361 annotated feedback--adjustment pairs, 95.6\% stay at the same merchant, versus 70.7\% among 916 pairs outside the surrounding island. Median intervals are 4.8 minutes versus 22.8 hours. Figure~\ref{fig:retry} follows one history from the region. C, D, and I identify anonymous cards, devices, and IP addresses; thought bubbles sketch the likely intent at each step. After a rejection, the next attempt uses a new card on the same device and IP address. After a second rejection, card and device both change while the IP address stays. The third attempt, 120 seconds after the first, succeeds. Each switch appears in the event tokens as a new bit pattern in the card or device field. The encoder never sees risk labels, merchant links, or payment outcomes; these were attached only for inspection. The region thus reflects structure learned from transaction histories alone. Appendix~\ref{app:case} specifies the selection rule.

\subsection{Online Deployment (RQ4)}
\label{sec:online}
\textbf{Serving.} Ant International deploys TRIDENT-Seq across multiple payment businesses; SME deployment began on June 8, 2026 (Figure~\ref{fig:overview}(c)). SME has the sparsest histories of our three businesses, and pretraining leads both offline metrics there with full labels (Table~\ref{tab:main}). The feature platform retrieves the available history, and the service applies the same event encoding as training~\cite{sculley2015}. The adapted encoder and risk head return a score. The decision system maps it to interventions under predefined rules for different traffic profiles. With quantization and Cython optimizations on an A10 GPU, model inference averages 46.66\,ms, with P95 of 55.86\,ms and P99 of 80.16\,ms. The inference budget is 120\,ms. History retrieval takes 50--60\,ms of the 250\,ms end-to-end budget.

\textbf{A/B test design.} At Ant International, we compared TRIDENT-Seq with XGB-AutoRefit, an iteratively refitted gradient-boosted tree model~\cite{chen2016xgboost}, and a SAINT-style deep tabular model~\cite{somepalli2021saint}. From June 8 to October 14, 2026, 30\% of platform transactions, about 28.20 million, were randomly assigned in equal shares to the three arms at launch~\cite{kohavi2007}. The amount-based chargeback (CB) rate divides chargeback amount by successful transaction amount. The overall loss rate uses financial losses after dispute-related adjustments over the same denominator. Portfolio rates pool recorded amounts across businesses and transaction dates within each arm. Traffic shares shifted briefly during load testing and staged rollout; assignment stayed random.

\textbf{SME evaluation.} The SME experiment separates scoring from intervention, because blocked transactions cannot produce chargebacks. In Stage~I (June 8--August 2), all arms scored without intervening; subsequent risk feedback measures recall among the 0.5\% highest-scoring transactions. In Stage~II (August 3--October 14), each arm made production decisions under the same business policies and a 0.5\% intervention budget.

\textbf{Outcome timing.} Financial outcomes in Tables~\ref{tab:ab} and~\ref{tab:datecheck} include reports recorded through October 25, 2026, giving SME Stage~II transactions 11--83 days of follow-up. SME estimates use 66 common transaction dates with CB amounts and positive successful amounts in all arms. Appendix~\ref{app:onlinecheck} reports historical label-count delays and paired calendar-week bootstrap intervals for the observed financial snapshot.

\textbf{Results.} Across the portfolio, TRIDENT-Seq records 37.74\% lower CB and 37.88\% lower overall loss rates than XGB-AutoRefit; reductions versus SAINT-style are 28.26\% and 24.07\%. All gaps exceed the business thresholds of 4\% for CB and 10\% for loss (Table~\ref{tab:ab}). In SME Stage~I, TRIDENT-Seq captures 29.79\% of risk transactions within the 0.5\% budget, versus 7.39\% for XGB-AutoRefit and 16.52\% for SAINT-style. In SME Stage~II, its pooled CB reduction versus XGB is 28.28\% (95\% CI: 10.69--42.38\%); common date weights give 26.09\% (9.24--41.34\%). SME Stage~I recall follows TRIDENT-Seq $>$ SAINT-style $>$ XGB-AutoRefit in all routinely monitored segments.

\textbf{Data governance.} Transaction data are used for authorized fraud prevention, with access restricted to authorized personnel in an internal environment. Online experiments follow existing risk policies, monitor approval rates and false declines, and support rollback. We report aggregate results and omit identifying details from illustrative cases.

\begin{table}[t]
\caption{Portfolio-level online A/B results at Ant International. Pooled arm-level rate changes relative to XGB.}
\label{tab:ab}
\small\setlength{\tabcolsep}{3pt}
\begin{tabular*}{\columnwidth}{@{\extracolsep{\fill}}lrrr@{}}
\toprule
Method & Traffic (M) & \makecell[r]{CB rate\\change $\downarrow$} & \makecell[r]{Loss rate\\change $\downarrow$}\\
\midrule
XGB-AutoRefit & 9.06 & base & base\\
SAINT-style & 10.56 & $-$13.21\% & $-$18.18\%\\
\textbf{TRIDENT-Seq} & 8.58 & \textbf{$-$37.74\%} & \textbf{$-$37.88\%}\\
\bottomrule
\end{tabular*}
\end{table}

\section{Conclusion}
TRIDENT-Seq pretrains a shared transaction encoder for payment risk with PFC. PFC selects future summaries with similar activity profiles, narrowing coarse activity differences among contrastive candidates. One pretrained encoder improves risk ranking across three payment businesses. It surpasses full-label scratch training in mean AP and recall with 10\% to 25\% of the training labels, and it transfers when the target business is excluded from pretraining.

The benefit depends on history coverage. On SME, 76.91\% of traffic has no history, and the encoder receives no events for it. Our results rest on one pretrained checkpoint, and Wallet AP varies considerably across fine-tuning runs. Natural next steps are to probe directly which cues the encoder relies on, to scale the pretraining data and model, and to combine sequence representations with production risk indicators. More broadly, our results suggest that the choice of contrastive candidates deserves as much attention as the prediction target.

\bibliographystyle{ACM-Reference-Format}
\bibliography{refs}


\begin{thebibliography}{50}


\ifx \showCODEN    \undefined \def \showCODEN     #1{\unskip}     \fi
\ifx \showISBNx    \undefined \def \showISBNx     #1{\unskip}     \fi
\ifx \showISBNxiii \undefined \def \showISBNxiii  #1{\unskip}     \fi
\ifx \showISSN     \undefined \def \showISSN      #1{\unskip}     \fi
\ifx \showLCCN     \undefined \def \showLCCN      #1{\unskip}     \fi
\ifx \shownote     \undefined \def \shownote      #1{#1}          \fi
\ifx \showarticletitle \undefined \def \showarticletitle #1{#1}   \fi
\ifx \showURL      \undefined \def \showURL       {\relax}        \fi
\providecommand\bibfield[2]{#2}
\providecommand\bibinfo[2]{#2}
\providecommand\natexlab[1]{#1}
\providecommand\showeprint[2][]{arXiv:#2}
\makeatletter
\@ifundefined{NAT@parse@date}{}{\let\NAT@parse@date@orig\NAT@parse@date}
\@ifundefined{NAT@parse@date}{}{\def\NAT@parse@date#1#2#3#4#5#6@@{\NAT@parse@date@orig#1#2#3#4#5#6@@\def\NAT@tempyear{0000}\def\NAT@tempexlab{{?}}\ifx\NAT@year\NAT@tempyear\ifx\NAT@exlab\NAT@tempexlab\def\NAT@date{[n.\,d.]}\else\edef\NAT@date{[n.\,d.]\NAT@exlab}\fi\fi}}
\makeatother

\bibitem[Affaki(2025)]%
        {stripe2025}
\bibfield{author}{\bibinfo{person}{John Affaki}.}
  \bibinfo{year}{2025}\natexlab{}.
\newblock \bibinfo{title}{Using {AI} to Optimize Payments Performance with the
  Payments Intelligence Suite}.
\newblock \bibinfo{howpublished}{Stripe Blog}.
\newblock
\urldef\tempurl%
\url{https://stripe.com/blog/using-ai-optimize-payments-performance-payments-intelligence-suite}
\showURL{%
\tempurl}


\bibitem[Babaev et~al\mbox{.}(2022)]%
        {babaev2022coles}
\bibfield{author}{\bibinfo{person}{Dmitrii Babaev} {et~al\mbox{.}}}
  \bibinfo{year}{2022}\natexlab{}.
\newblock \showarticletitle{{CoLES}: Contrastive Learning for Event Sequences
  with Self-Supervision}. In \bibinfo{booktitle}{\emph{Proceedings of the 2022
  International Conference on Management of Data}}. \bibinfo{publisher}{ACM},
  \bibinfo{address}{New York, NY, USA}, \bibinfo{pages}{1190--1199}.
\newblock
\href{https://doi.org/10.1145/3514221.3526129}{doi:\nolinkurl{10.1145/3514221.3526129}}


\bibitem[Bahnsen et~al\mbox{.}(2016)]%
        {bahnsen2016}
\bibfield{author}{\bibinfo{person}{Alejandro~Correa Bahnsen},
  \bibinfo{person}{Djamila Aouada}, \bibinfo{person}{Aleksandar Stojanovic},
  {and} \bibinfo{person}{Bj{\"o}rn Ottersten}.}
  \bibinfo{year}{2016}\natexlab{}.
\newblock \showarticletitle{Feature Engineering Strategies for Credit Card
  Fraud Detection}.
\newblock \bibinfo{journal}{\emph{Expert Systems with Applications}}
  \bibinfo{volume}{51} (\bibinfo{year}{2016}), \bibinfo{pages}{134--142}.
\newblock


\bibitem[Bardes et~al\mbox{.}(2022)]%
        {bardes2022vicreg}
\bibfield{author}{\bibinfo{person}{Adrien Bardes}, \bibinfo{person}{Jean
  Ponce}, {and} \bibinfo{person}{Yann LeCun}.} \bibinfo{year}{2022}\natexlab{}.
\newblock \showarticletitle{{VICReg}: Variance-Invariance-Covariance
  Regularization for Self-Supervised Learning}. In
  \bibinfo{booktitle}{\emph{ICLR}}.
\newblock


\bibitem[Bloom(1970)]%
        {bloom1970}
\bibfield{author}{\bibinfo{person}{Burton~H. Bloom}.}
  \bibinfo{year}{1970}\natexlab{}.
\newblock \showarticletitle{Space/Time Trade-offs in Hash Coding with Allowable
  Errors}.
\newblock \bibinfo{journal}{\emph{Commun. ACM}} \bibinfo{volume}{13},
  \bibinfo{number}{7} (\bibinfo{year}{1970}), \bibinfo{pages}{422--426}.
\newblock


\bibitem[Carcillo et~al\mbox{.}(2018)]%
        {carcillo2018scarff}
\bibfield{author}{\bibinfo{person}{Fabrizio Carcillo}, \bibinfo{person}{Andrea
  Dal~Pozzolo}, \bibinfo{person}{Yann-A{\"e}l Le~Borgne},
  \bibinfo{person}{Olivier Caelen}, \bibinfo{person}{Yannis Mazzer}, {and}
  \bibinfo{person}{Gianluca Bontempi}.} \bibinfo{year}{2018}\natexlab{}.
\newblock \showarticletitle{{SCARFF}: A Scalable Framework for Streaming Credit
  Card Fraud Detection with {Spark}}.
\newblock \bibinfo{journal}{\emph{Information Fusion}}  \bibinfo{volume}{41}
  (\bibinfo{year}{2018}), \bibinfo{pages}{182--194}.
\newblock


\bibitem[Chen et~al\mbox{.}(2024)]%
        {chen2024ccfd}
\bibfield{author}{\bibinfo{person}{Chiao-Ting Chen}, \bibinfo{person}{Chi Lee},
  \bibinfo{person}{Szu-Hao Huang}, {and} \bibinfo{person}{Wen-Chih Peng}.}
  \bibinfo{year}{2024}\natexlab{}.
\newblock \showarticletitle{Credit Card Fraud Detection via Intelligent
  Sampling and Self-supervised Learning}.
\newblock \bibinfo{journal}{\emph{ACM Transactions on Intelligent Systems and
  Technology}} \bibinfo{volume}{15}, \bibinfo{number}{2}, Article
  \bibinfo{articleno}{35} (\bibinfo{year}{2024}), \bibinfo{numpages}{29}~pages.
\newblock
\href{https://doi.org/10.1145/3641283}{doi:\nolinkurl{10.1145/3641283}}


\bibitem[Chen and Guestrin(2016)]%
        {chen2016xgboost}
\bibfield{author}{\bibinfo{person}{Tianqi Chen} {and} \bibinfo{person}{Carlos
  Guestrin}.} \bibinfo{year}{2016}\natexlab{}.
\newblock \showarticletitle{{XGBoost}: A Scalable Tree Boosting System}. In
  \bibinfo{booktitle}{\emph{KDD}}. \bibinfo{pages}{785--794}.
\newblock


\bibitem[Chen et~al\mbox{.}(2020)]%
        {chen2020simclr}
\bibfield{author}{\bibinfo{person}{Ting Chen}, \bibinfo{person}{Simon
  Kornblith}, \bibinfo{person}{Mohammad Norouzi}, {and}
  \bibinfo{person}{Geoffrey Hinton}.} \bibinfo{year}{2020}\natexlab{}.
\newblock \showarticletitle{A Simple Framework for Contrastive Learning of
  Visual Representations}. In \bibinfo{booktitle}{\emph{ICML}}.
  \bibinfo{pages}{1597--1607}.
\newblock


\bibitem[Chetlapalli and Yao(2026)]%
        {plaid2026a}
\bibfield{author}{\bibinfo{person}{Raghu Chetlapalli} {and}
  \bibinfo{person}{Wen Yao}.} \bibinfo{year}{2026}\natexlab{}.
\newblock \bibinfo{title}{Learning the Grammar of Money with a Sequential
  Foundation Model}.
\newblock \bibinfo{howpublished}{Plaid Engineering Blog}.
\newblock
\urldef\tempurl%
\url{https://plaid.com/blog/sequential-foundation-model/}
\showURL{%
\tempurl}


\bibitem[Dal~Pozzolo et~al\mbox{.}(2018)]%
        {dalpozzolo2018}
\bibfield{author}{\bibinfo{person}{Andrea Dal~Pozzolo} {et~al\mbox{.}}}
  \bibinfo{year}{2018}\natexlab{}.
\newblock \showarticletitle{Credit Card Fraud Detection: A Realistic Modeling
  and a Novel Learning Strategy}.
\newblock \bibinfo{journal}{\emph{IEEE Transactions on Neural Networks and
  Learning Systems}} \bibinfo{volume}{29}, \bibinfo{number}{8}
  (\bibinfo{year}{2018}), \bibinfo{pages}{3784--3797}.
\newblock


\bibitem[Davis and Goadrich(2006)]%
        {davis2006pr}
\bibfield{author}{\bibinfo{person}{Jesse Davis} {and} \bibinfo{person}{Mark
  Goadrich}.} \bibinfo{year}{2006}\natexlab{}.
\newblock \showarticletitle{The Relationship Between Precision-Recall and {ROC}
  Curves}. In \bibinfo{booktitle}{\emph{ICML}}. \bibinfo{pages}{233--240}.
\newblock


\bibitem[Dou et~al\mbox{.}(2026)]%
        {dou2025tgpt}
\bibfield{author}{\bibinfo{person}{Yingtong Dou} {et~al\mbox{.}}}
  \bibinfo{year}{2026}\natexlab{}.
\newblock \showarticletitle{{TransactionGPT}: Toward Foundational Transaction
  Modeling}. In \bibinfo{booktitle}{\emph{Proceedings of the 32nd ACM SIGKDD
  Conference on Knowledge Discovery and Data Mining V.2}}.
  \bibinfo{publisher}{ACM}, \bibinfo{address}{New York, NY, USA},
  \bibinfo{pages}{7185--7196}.
\newblock
\href{https://doi.org/10.1145/3770855.3818496}{doi:\nolinkurl{10.1145/3770855.3818496}}


\bibitem[Dou et~al\mbox{.}(2020)]%
        {dou2020care}
\bibfield{author}{\bibinfo{person}{Yingtong Dou}, \bibinfo{person}{Zhiwei Liu},
  \bibinfo{person}{Li Sun}, \bibinfo{person}{Yutong Deng}, \bibinfo{person}{Hao
  Peng}, {and} \bibinfo{person}{Philip~S. Yu}.}
  \bibinfo{year}{2020}\natexlab{}.
\newblock \showarticletitle{Enhancing Graph Neural Network-based Fraud
  Detectors against Camouflaged Fraudsters}. In
  \bibinfo{booktitle}{\emph{CIKM}}. \bibinfo{pages}{315--324}.
\newblock


\bibitem[Fu et~al\mbox{.}(2025)]%
        {fu2025ctr}
\bibfield{author}{\bibinfo{person}{Zichuan Fu}, \bibinfo{person}{Xiangyang Li},
  \bibinfo{person}{Chuhan Wu}, \bibinfo{person}{Yichao Wang},
  \bibinfo{person}{Kuicai Dong}, \bibinfo{person}{Xiangyu Zhao},
  \bibinfo{person}{Mengchen Zhao}, \bibinfo{person}{Huifeng Guo}, {and}
  \bibinfo{person}{Ruiming Tang}.} \bibinfo{year}{2025}\natexlab{}.
\newblock \showarticletitle{A Unified Framework for Multi-Domain {CTR}
  Prediction via Large Language Models}.
\newblock \bibinfo{journal}{\emph{ACM Transactions on Information Systems}}
  \bibinfo{volume}{43}, \bibinfo{number}{5}, Article \bibinfo{articleno}{117}
  (\bibinfo{year}{2025}).
\newblock


\bibitem[Gong et~al\mbox{.}(2023)]%
        {gong2023coldstart}
\bibfield{author}{\bibinfo{person}{Yuqi Gong} {et~al\mbox{.}}}
  \bibinfo{year}{2023}\natexlab{}.
\newblock \bibinfo{title}{An Unified Search and Recommendation Foundation Model
  for Cold-Start Scenario}.
\newblock \bibinfo{howpublished}{arXiv:2309.08939}.
\newblock


\bibitem[Grill et~al\mbox{.}(2020)]%
        {grill2020byol}
\bibfield{author}{\bibinfo{person}{Jean-Bastien Grill} {et~al\mbox{.}}}
  \bibinfo{year}{2020}\natexlab{}.
\newblock \showarticletitle{Bootstrap Your Own Latent: A New Approach to
  Self-Supervised Learning}. In \bibinfo{booktitle}{\emph{NeurIPS}}.
\newblock


\bibitem[Hochreiter and Schmidhuber(1997)]%
        {hochreiter1997lstm}
\bibfield{author}{\bibinfo{person}{Sepp Hochreiter} {and}
  \bibinfo{person}{J{\"u}rgen Schmidhuber}.} \bibinfo{year}{1997}\natexlab{}.
\newblock \showarticletitle{Long Short-Term Memory}.
\newblock \bibinfo{journal}{\emph{Neural Computation}} \bibinfo{volume}{9},
  \bibinfo{number}{8} (\bibinfo{year}{1997}), \bibinfo{pages}{1735--1780}.
\newblock


\bibitem[Horvitz and Thompson(1952)]%
        {horvitz1952}
\bibfield{author}{\bibinfo{person}{D.~G. Horvitz} {and} \bibinfo{person}{D.~J.
  Thompson}.} \bibinfo{year}{1952}\natexlab{}.
\newblock \showarticletitle{A Generalization of Sampling Without Replacement
  from a Finite Universe}.
\newblock \bibinfo{journal}{\emph{J. Amer. Statist. Assoc.}}
  \bibinfo{volume}{47}, \bibinfo{number}{260} (\bibinfo{year}{1952}),
  \bibinfo{pages}{663--685}.
\newblock


\bibitem[Hou et~al\mbox{.}(2022)]%
        {hou2022unisrec}
\bibfield{author}{\bibinfo{person}{Yupeng Hou}, \bibinfo{person}{Shanlei Mu},
  \bibinfo{person}{Wayne~Xin Zhao}, \bibinfo{person}{Yaliang Li},
  \bibinfo{person}{Bolin Ding}, {and} \bibinfo{person}{Ji-Rong Wen}.}
  \bibinfo{year}{2022}\natexlab{}.
\newblock \showarticletitle{Towards Universal Sequence Representation Learning
  for Recommender Systems}. In \bibinfo{booktitle}{\emph{Proceedings of the
  28th ACM SIGKDD Conference on Knowledge Discovery and Data Mining}}.
  \bibinfo{publisher}{ACM}, \bibinfo{address}{New York, NY, USA},
  \bibinfo{pages}{585--593}.
\newblock
\href{https://doi.org/10.1145/3534678.3539381}{doi:\nolinkurl{10.1145/3534678.3539381}}


\bibitem[Jin(2026)]%
        {plaid2026b}
\bibfield{author}{\bibinfo{person}{Chris Jin}.}
  \bibinfo{year}{2026}\natexlab{}.
\newblock \bibinfo{title}{Fraud Foundation Model: The Next Chapter in Fraud
  Prevention for Plaid}.
\newblock \bibinfo{howpublished}{Plaid Blog}.
\newblock
\urldef\tempurl%
\url{https://plaid.com/blog/introducing-fraud-foundation-model-for-protect/}
\showURL{%
\tempurl}


\bibitem[Jurgovsky et~al\mbox{.}(2018)]%
        {jurgovsky2018}
\bibfield{author}{\bibinfo{person}{Johannes Jurgovsky} {et~al\mbox{.}}}
  \bibinfo{year}{2018}\natexlab{}.
\newblock \showarticletitle{Sequence Classification for Credit-Card Fraud
  Detection}.
\newblock \bibinfo{journal}{\emph{Expert Systems with Applications}}
  \bibinfo{volume}{100} (\bibinfo{year}{2018}), \bibinfo{pages}{234--245}.
\newblock


\bibitem[Kazemi et~al\mbox{.}(2019)]%
        {kazemi2019time2vec}
\bibfield{author}{\bibinfo{person}{Seyed~Mehran Kazemi} {et~al\mbox{.}}}
  \bibinfo{year}{2019}\natexlab{}.
\newblock \bibinfo{title}{{Time2Vec}: Learning a Vector Representation of
  Time}.
\newblock \bibinfo{howpublished}{arXiv:1907.05321}.
\newblock


\bibitem[Keilwagen et~al\mbox{.}(2014)]%
        {keilwagen2014}
\bibfield{author}{\bibinfo{person}{Jens Keilwagen}, \bibinfo{person}{Ivo
  Grosse}, {and} \bibinfo{person}{Jan Grau}.} \bibinfo{year}{2014}\natexlab{}.
\newblock \showarticletitle{Area under Precision-Recall Curves for Weighted and
  Unweighted Data}.
\newblock \bibinfo{journal}{\emph{PLOS ONE}} \bibinfo{volume}{9},
  \bibinfo{number}{3} (\bibinfo{year}{2014}), \bibinfo{pages}{e92209}.
\newblock


\bibitem[Kohavi et~al\mbox{.}(2007)]%
        {kohavi2007}
\bibfield{author}{\bibinfo{person}{Ron Kohavi}, \bibinfo{person}{Randal~M.
  Henne}, {and} \bibinfo{person}{Dan Sommerfield}.}
  \bibinfo{year}{2007}\natexlab{}.
\newblock \showarticletitle{Practical Guide to Controlled Experiments on the
  Web: Listen to Your Customers not to the {HiPPO}}. In
  \bibinfo{booktitle}{\emph{KDD}}. \bibinfo{pages}{959--967}.
\newblock


\bibitem[Li et~al\mbox{.}(2025)]%
        {li2025panther}
\bibfield{author}{\bibinfo{person}{Guilin Li} {et~al\mbox{.}}}
  \bibinfo{year}{2025}\natexlab{}.
\newblock \showarticletitle{{PANTHER}: Generative Pretraining Beyond Language
  for Sequential User Behavior Modeling}. In \bibinfo{booktitle}{\emph{Advances
  in Neural Information Processing Systems}}, Vol.~\bibinfo{volume}{38}.
  \bibinfo{publisher}{Neural Information Processing Systems Foundation},
  \bibinfo{pages}{181532--181553}.
\newblock
\href{https://doi.org/10.52202/085713-5462}{doi:\nolinkurl{10.52202/085713-5462}}


\bibitem[Liang et~al\mbox{.}(2025)]%
        {liang2025exfm}
\bibfield{author}{\bibinfo{person}{Mingfu Liang} {et~al\mbox{.}}}
  \bibinfo{year}{2025}\natexlab{}.
\newblock \showarticletitle{External Large Foundation Model: How to Efficiently
  Serve Trillions of Parameters for Online Ads Recommendation}. In
  \bibinfo{booktitle}{\emph{Companion Proceedings of the ACM on Web Conference
  2025}}. \bibinfo{publisher}{ACM}, \bibinfo{address}{New York, NY, USA},
  \bibinfo{pages}{344--353}.
\newblock
\href{https://doi.org/10.1145/3701716.3715223}{doi:\nolinkurl{10.1145/3701716.3715223}}


\bibitem[Liu et~al\mbox{.}(2021)]%
        {liu2021pcgnn}
\bibfield{author}{\bibinfo{person}{Yang Liu}, \bibinfo{person}{Xiang Ao},
  \bibinfo{person}{Zidi Qin}, \bibinfo{person}{Jianfeng Chi},
  \bibinfo{person}{Jinghua Feng}, \bibinfo{person}{Hao Yang}, {and}
  \bibinfo{person}{Qing He}.} \bibinfo{year}{2021}\natexlab{}.
\newblock \showarticletitle{Pick and Choose: A {GNN}-based Imbalanced Learning
  Approach for Fraud Detection}. In \bibinfo{booktitle}{\emph{WWW}}.
  \bibinfo{pages}{3168--3177}.
\newblock


\bibitem[Liu et~al\mbox{.}(2018)]%
        {liu2018gem}
\bibfield{author}{\bibinfo{person}{Ziqi Liu}, \bibinfo{person}{Chaochao Chen},
  \bibinfo{person}{Xinxing Yang}, \bibinfo{person}{Jun Zhou},
  \bibinfo{person}{Xiaolong Li}, {and} \bibinfo{person}{Le Song}.}
  \bibinfo{year}{2018}\natexlab{}.
\newblock \showarticletitle{Heterogeneous Graph Neural Networks for Malicious
  Account Detection}. In \bibinfo{booktitle}{\emph{CIKM}}.
  \bibinfo{pages}{2077--2085}.
\newblock


\bibitem[Loshchilov and Hutter(2017)]%
        {loshchilov2017sgdr}
\bibfield{author}{\bibinfo{person}{Ilya Loshchilov} {and}
  \bibinfo{person}{Frank Hutter}.} \bibinfo{year}{2017}\natexlab{}.
\newblock \showarticletitle{{SGDR}: Stochastic Gradient Descent with Warm
  Restarts}. In \bibinfo{booktitle}{\emph{ICLR}}.
\newblock


\bibitem[Loshchilov and Hutter(2019)]%
        {loshchilov2019adamw}
\bibfield{author}{\bibinfo{person}{Ilya Loshchilov} {and}
  \bibinfo{person}{Frank Hutter}.} \bibinfo{year}{2019}\natexlab{}.
\newblock \showarticletitle{Decoupled Weight Decay Regularization}. In
  \bibinfo{booktitle}{\emph{ICLR}}.
\newblock


\bibitem[Ostroukhov et~al\mbox{.}(2026)]%
        {pragma2026}
\bibfield{author}{\bibinfo{person}{Maxim Ostroukhov} {et~al\mbox{.}}}
  \bibinfo{year}{2026}\natexlab{}.
\newblock \bibinfo{title}{{PRAGMA}: Revolut Foundation Model}.
\newblock \bibinfo{howpublished}{arXiv:2604.08649}.
\newblock
\href{https://doi.org/10.48550/arXiv.2604.08649}{doi:\nolinkurl{10.48550/arXiv.2604.08649}}


\bibitem[Padhi et~al\mbox{.}(2021)]%
        {padhi2021tabbert}
\bibfield{author}{\bibinfo{person}{Inkit Padhi} {et~al\mbox{.}}}
  \bibinfo{year}{2021}\natexlab{}.
\newblock \showarticletitle{Tabular Transformers for Modeling Multivariate Time
  Series}. In \bibinfo{booktitle}{\emph{ICASSP}}. \bibinfo{pages}{3565--3569}.
\newblock


\bibitem[Pan and Yang(2010)]%
        {pan2010transfer}
\bibfield{author}{\bibinfo{person}{Sinno~Jialin Pan} {and}
  \bibinfo{person}{Qiang Yang}.} \bibinfo{year}{2010}\natexlab{}.
\newblock \showarticletitle{A Survey on Transfer Learning}.
\newblock \bibinfo{journal}{\emph{IEEE TKDE}} \bibinfo{volume}{22},
  \bibinfo{number}{10} (\bibinfo{year}{2010}), \bibinfo{pages}{1345--1359}.
\newblock


\bibitem[Robinson et~al\mbox{.}(2021a)]%
        {robinson2021hard}
\bibfield{author}{\bibinfo{person}{Joshua Robinson}, \bibinfo{person}{Ching-Yao
  Chuang}, \bibinfo{person}{Suvrit Sra}, {and} \bibinfo{person}{Stefanie
  Jegelka}.} \bibinfo{year}{2021}\natexlab{a}.
\newblock \showarticletitle{Contrastive Learning with Hard Negative Samples}.
  In \bibinfo{booktitle}{\emph{ICLR}}.
\newblock


\bibitem[Robinson et~al\mbox{.}(2021b)]%
        {robinson2021shortcut}
\bibfield{author}{\bibinfo{person}{Joshua~W. Robinson}, \bibinfo{person}{Li
  Sun}, \bibinfo{person}{Ke Yu}, \bibinfo{person}{Kayhan Batmanghelich},
  \bibinfo{person}{Stefanie Jegelka}, {and} \bibinfo{person}{Suvrit Sra}.}
  \bibinfo{year}{2021}\natexlab{b}.
\newblock \showarticletitle{Can Contrastive Learning Avoid Shortcut
  Solutions?}. In \bibinfo{booktitle}{\emph{NeurIPS}}.
\newblock


\bibitem[Saito and Rehmsmeier(2015)]%
        {saito2015pr}
\bibfield{author}{\bibinfo{person}{Takaya Saito} {and} \bibinfo{person}{Marc
  Rehmsmeier}.} \bibinfo{year}{2015}\natexlab{}.
\newblock \showarticletitle{The Precision-Recall Plot Is More Informative than
  the {ROC} Plot When Evaluating Binary Classifiers on Imbalanced Datasets}.
\newblock \bibinfo{journal}{\emph{PLOS ONE}} \bibinfo{volume}{10},
  \bibinfo{number}{3} (\bibinfo{year}{2015}), \bibinfo{pages}{e0118432}.
\newblock


\bibitem[Sculley et~al\mbox{.}(2015)]%
        {sculley2015}
\bibfield{author}{\bibinfo{person}{D. Sculley} {et~al\mbox{.}}}
  \bibinfo{year}{2015}\natexlab{}.
\newblock \showarticletitle{Hidden Technical Debt in Machine Learning Systems}.
  In \bibinfo{booktitle}{\emph{Advances in Neural Information Processing
  Systems}}, Vol.~\bibinfo{volume}{28}. \bibinfo{pages}{2503--2511}.
\newblock


\bibitem[Senuma(2025)]%
        {senuma2025mmh3}
\bibfield{author}{\bibinfo{person}{Hajime Senuma}.}
  \bibinfo{year}{2025}\natexlab{}.
\newblock \showarticletitle{{mmh3}: A {Python} extension for {MurmurHash3}}.
\newblock \bibinfo{journal}{\emph{Journal of Open Source Software}}
  \bibinfo{volume}{10}, \bibinfo{number}{105} (\bibinfo{year}{2025}),
  \bibinfo{pages}{6124}.
\newblock
\href{https://doi.org/10.21105/joss.06124}{doi:\nolinkurl{10.21105/joss.06124}}


\bibitem[Skalski et~al\mbox{.}(2023)]%
        {skalski2023fpm}
\bibfield{author}{\bibinfo{person}{Piotr Skalski} {et~al\mbox{.}}}
  \bibinfo{year}{2023}\natexlab{}.
\newblock \showarticletitle{Towards a Foundation Purchasing Model: Pretrained
  Generative Autoregression on Transaction Sequences}. In
  \bibinfo{booktitle}{\emph{Proceedings of the Fourth ACM International
  Conference on AI in Finance}}. \bibinfo{publisher}{ACM},
  \bibinfo{address}{New York, NY, USA}, \bibinfo{pages}{141--149}.
\newblock
\href{https://doi.org/10.1145/3604237.3626850}{doi:\nolinkurl{10.1145/3604237.3626850}}


\bibitem[Somepalli et~al\mbox{.}(2021)]%
        {somepalli2021saint}
\bibfield{author}{\bibinfo{person}{Gowthami Somepalli}, \bibinfo{person}{Micah
  Goldblum}, \bibinfo{person}{Avi Schwarzschild}, \bibinfo{person}{C.~Bayan
  Bruss}, {and} \bibinfo{person}{Tom Goldstein}.}
  \bibinfo{year}{2021}\natexlab{}.
\newblock \bibinfo{title}{{SAINT}: Improved Neural Networks for Tabular Data
  via Row Attention and Contrastive Pre-Training}.
\newblock \bibinfo{howpublished}{arXiv:2106.01342}.
\newblock


\bibitem[Svenstrup et~al\mbox{.}(2017)]%
        {svenstrup2017hash}
\bibfield{author}{\bibinfo{person}{Dan Svenstrup},
  \bibinfo{person}{Jonas~Meinertz Hansen}, {and} \bibinfo{person}{Ole
  Winther}.} \bibinfo{year}{2017}\natexlab{}.
\newblock \showarticletitle{Hash Embeddings for Efficient Word
  Representations}. In \bibinfo{booktitle}{\emph{Advances in Neural Information
  Processing Systems}}, Vol.~\bibinfo{volume}{30}. \bibinfo{pages}{4928--4936}.
\newblock


\bibitem[van~den Oord et~al\mbox{.}(2018)]%
        {oord2018cpc}
\bibfield{author}{\bibinfo{person}{Aaron van~den Oord}, \bibinfo{person}{Yazhe
  Li}, {and} \bibinfo{person}{Oriol Vinyals}.} \bibinfo{year}{2018}\natexlab{}.
\newblock \bibinfo{title}{Representation Learning with Contrastive Predictive
  Coding}.
\newblock \bibinfo{howpublished}{arXiv:1807.03748}.
\newblock


\bibitem[van~der Maaten and Hinton(2008)]%
        {maaten2008tsne}
\bibfield{author}{\bibinfo{person}{Laurens van~der Maaten} {and}
  \bibinfo{person}{Geoffrey Hinton}.} \bibinfo{year}{2008}\natexlab{}.
\newblock \showarticletitle{Visualizing Data using t-{SNE}}.
\newblock \bibinfo{journal}{\emph{JMLR}}  \bibinfo{volume}{9}
  (\bibinfo{year}{2008}), \bibinfo{pages}{2579--2605}.
\newblock


\bibitem[Vaswani et~al\mbox{.}(2017)]%
        {vaswani2017}
\bibfield{author}{\bibinfo{person}{Ashish Vaswani}, \bibinfo{person}{Noam
  Shazeer}, \bibinfo{person}{Niki Parmar}, \bibinfo{person}{Jakob Uszkoreit},
  \bibinfo{person}{Llion Jones}, \bibinfo{person}{Aidan~N. Gomez},
  \bibinfo{person}{{\L}ukasz Kaiser}, {and} \bibinfo{person}{Illia
  Polosukhin}.} \bibinfo{year}{2017}\natexlab{}.
\newblock \showarticletitle{Attention Is All You Need}. In
  \bibinfo{booktitle}{\emph{Advances in Neural Information Processing
  Systems}}, Vol.~\bibinfo{volume}{30}. \bibinfo{pages}{5998--6008}.
\newblock


\bibitem[Wang and Isola(2020)]%
        {wang2020align}
\bibfield{author}{\bibinfo{person}{Tongzhou Wang} {and}
  \bibinfo{person}{Phillip Isola}.} \bibinfo{year}{2020}\natexlab{}.
\newblock \showarticletitle{Understanding Contrastive Representation Learning
  through Alignment and Uniformity on the Hypersphere}. In
  \bibinfo{booktitle}{\emph{ICML}}. \bibinfo{pages}{9929--9939}.
\newblock


\bibitem[Weinberger et~al\mbox{.}(2009)]%
        {weinberger2009hashing}
\bibfield{author}{\bibinfo{person}{Kilian Weinberger}, \bibinfo{person}{Anirban
  Dasgupta}, \bibinfo{person}{John Langford}, \bibinfo{person}{Alex Smola},
  {and} \bibinfo{person}{Josh Attenberg}.} \bibinfo{year}{2009}\natexlab{}.
\newblock \showarticletitle{Feature Hashing for Large Scale Multitask
  Learning}. In \bibinfo{booktitle}{\emph{ICML}}. \bibinfo{pages}{1113--1120}.
\newblock


\bibitem[Wu et~al\mbox{.}(2025)]%
        {wu2025byb}
\bibfield{author}{\bibinfo{person}{Weichang Wu}, \bibinfo{person}{Xiaolu
  Zhang}, \bibinfo{person}{Jun Zhou}, \bibinfo{person}{Yuchen Li}, {and}
  \bibinfo{person}{Wenwen Xia}.} \bibinfo{year}{2025}\natexlab{}.
\newblock \bibinfo{title}{Bootstrapping Your Behavior: A New Pretraining
  Strategy for User Behavior Sequence Data}.
\newblock \bibinfo{howpublished}{arXiv:2506.11053}.
\newblock
\href{https://doi.org/10.48550/arXiv.2506.11053}{doi:\nolinkurl{10.48550/arXiv.2506.11053}}


\bibitem[Yue et~al\mbox{.}(2022)]%
        {yue2022ts2vec}
\bibfield{author}{\bibinfo{person}{Zhihan Yue}, \bibinfo{person}{Yujing Wang},
  \bibinfo{person}{Juanyong Duan}, \bibinfo{person}{Tianmeng Yang},
  \bibinfo{person}{Congrui Huang}, \bibinfo{person}{Yunhai Tong}, {and}
  \bibinfo{person}{Bixiong Xu}.} \bibinfo{year}{2022}\natexlab{}.
\newblock \showarticletitle{{TS2Vec}: Towards Universal Representation of Time
  Series}. In \bibinfo{booktitle}{\emph{AAAI}}. \bibinfo{pages}{8980--8987}.
\newblock


\bibitem[Zhang et~al\mbox{.}(2026)]%
        {zhang2026friskgpt}
\bibfield{author}{\bibinfo{person}{Zhongjian Zhang}, \bibinfo{person}{Mengmei
  Zhang}, \bibinfo{person}{Dehua Xu}, \bibinfo{person}{Rongjun Shi},
  \bibinfo{person}{Jianfeng Liu}, \bibinfo{person}{Fuli Meng},
  \bibinfo{person}{Huajian Xu}, \bibinfo{person}{Xiao Wang},
  \bibinfo{person}{Ruijia Wang}, \bibinfo{person}{Junze Chen},
  \bibinfo{person}{Minwei Tang}, {and} \bibinfo{person}{Chuan Shi}.}
  \bibinfo{year}{2026}\natexlab{}.
\newblock \showarticletitle{{FRiskGPT}: A Generative Foundation Model for
  Financial Risk Detection}. In \bibinfo{booktitle}{\emph{Proceedings of the
  ACM Web Conference 2026}}. \bibinfo{publisher}{ACM}, \bibinfo{address}{New
  York, NY, USA}, \bibinfo{pages}{7733--7744}.
\newblock
\href{https://doi.org/10.1145/3774904.3792832}{doi:\nolinkurl{10.1145/3774904.3792832}}


\end{thebibliography}
\appendix
\raggedbottom
\section{Implementation Details}
\label{app:impl}
This appendix specifies event encoding, pretraining targets, and optimization. Table~\ref{tab:pre} summarizes encoder and pretraining settings, Table~\ref{tab:summary} defines the future-summary coordinates, and Table~\ref{tab:down} lists downstream optimization settings.

\subsection{Event Encoding and Network Configuration}
\textbf{Event encoding.} For each field value, MurmurHash3~\cite{senuma2025mmh3} maps the string \mbox{\texttt{field-name:value}} to a 128-bit hash with four 32-bit slices $s_0,\ldots,s_3$. Probe $k\in\{0,1,2,3\}$ sets bit $o_k=32k+(s_k\bmod32)$ of the 128-bit field vector, so each probe addresses its own 32-bit block and each nonempty value sets one bit per block. Missing values produce an all-zero vector. For example, block-local indices $(3,7,12,2)$ set positions $(3,39,76,98)$.

\textbf{Prediction heads.} The prefix head uses layer normalization and a $1024\to1024$ linear layer. The future head uses layer normalization and $24\to1024\to1024$ linear layers with GELU between them. The activity head uses $1024\to1024\to25$ layers with GELU. The risk head uses layer normalization, a 256-unit GELU layer, dropout 0.1, and a scalar output.

\begin{table}[htbp]\small
\caption{Encoder and pretraining configuration.}\label{tab:pre}
\begin{tabularx}{\columnwidth}{@{}>{\raggedright\arraybackslash}p{.42\columnwidth}>{\raggedright\arraybackslash}X@{}}\toprule
Setting & Value\\\midrule
Layers / hidden size / heads & 8 / 1,024 / 8\\
Encoder parameters / history cap & 92.92M / 300 events\\
Hash bits / probes & 128 / 4 per field\\
Eligible history / prefix & At least 2 events / older 60\%\\
Future summary & 24 dimensions (Table~\ref{tab:summary})\\
Profile & Three coordinates (Equation~\ref{eq:profile})\\
Eligibility / temperature & L1 distance $\leq1.5$ / 0.2\\
Activity weight & $\lambda_{\mathrm{act}}=0.30$\\
Dispersion weight / threshold & $\lambda_{\mathrm{disp}}=0.05$ / $\sigma_0=0.02$\\
Optimizer / learning rate & AdamW / $1.5\times10^{-4}$\\
Updates / global batch & 24,000 / 128\\
\bottomrule
\end{tabularx}\end{table}

\subsection{Pretraining Targets}
\label{app:targets}
\textbf{Future summary.} A history of $n\geq2$ events has a prefix of $\lfloor0.6n\rfloor$ events, clipped to $[1,n-1]$. For a future of $m$ events, amounts enter through $u_j=\log(1+\max(a_j,0))$ as in Equation~\ref{eq:profile}, and we let $\mathbf h_j=[\mathbf h_{j,1}\|\cdots\|\mathbf h_{j,12}]$ concatenate the twelve 128-bit field hashes of event $j$. Table~\ref{tab:summary} defines the 24 coordinates; histogram counts are normalized by $m$. Moments use population standard deviations and exclude padding; amounts and ages are clipped before these calculations. The switching rate of $m$ events is the fraction of field values that change between adjacent events,
\begin{equation}
\mathrm{SW}=\frac{1}{12(m-1)}\sum_{j=2}^{m}\sum_{k=1}^{12}\mathbf{1}[\mathbf h_{j,k}\ne\mathbf h_{j-1,k}].
\end{equation}
Adjacent-event statistics are zero for a one-event future. Let $\Delta\delta$ contain consecutive valid age differences.

\begin{table}[htbp]
\caption{Future-summary coordinates.}
\label{tab:summary}
\small\setlength{\tabcolsep}{3pt}
\begin{tabularx}{\columnwidth}{@{}lX@{}}\toprule
Coordinates & Definition\\\midrule
1 & Constant 1\\
2 & $\min(10,\operatorname{Std}(\Delta\delta)/\max(\operatorname{Mean}|\Delta\delta|,10^{-6}))/10$\\
3--5 & $\operatorname{Mean}(u)/16$, $\operatorname{Std}(u)/16$, $\max(u)/16$\\
6--7 & $\operatorname{Mean}(\delta)/92$, $\operatorname{Std}(\delta)/92$\\
8 & $\sum_{j,k}h_{jk}/(1536m)$\\
9 & Field-level switching rate SW between adjacent events\\
10 & Number of occupied bits in $\bigvee_j \mathbf h_j$, divided by 1,536\\
11--18 & Eight equal-width histogram bins of $u$ on $[0,16]$\\
19--24 & Six equal-width histogram bins of $\delta$ on $[0,92]$\\\bottomrule
\end{tabularx}\end{table}

\textbf{Activity target.} For each window width $w\in\{75,150,300\}$, the window holds the $w$ most recent events of the full training history, or all events if fewer. Let $n_w$ count its events, $U_w$ be its union hash occupancy divided by 1,536, and $C_w$ its switching rate SW. The six window statistics are
\begin{equation}
K_w=\operatorname{clip}_{[0,1]}\!\left[\frac{n_w}{w},\frac{\log(1+\sum u_j)}{16},\frac{\operatorname{Mean}(u)}{16},\frac{\operatorname{Mean}(\delta)}{92},U_w,C_w\right].
\end{equation}
The 25-dimensional target $A(H)$ concatenates $K_{75}$, $K_{150}$, $K_{300}$, coordinates 2, 4, 5, and 6 of $K_{300}-K_{75}$, which contrast the most recent window with the longer history, and three zeros. Empty-window and undefined statistics are zero. Smooth L1 uses $\beta=1$ and averages over the batch and the 25 coordinates.

\subsection{Optimization and Protocol}
\textbf{Ablation variants.} Random negatives retain each row's matched-candidate count, including the fallback count. Shuffled pairs apply a one-position cyclic shift to the future summaries and their profiles before the mask is constructed.

\textbf{Downstream adaptation.} The encoder and risk head are fine-tuned jointly with class-balanced binary cross-entropy, with class weights computed within each training-label subset. Weighted validation AP selects learning rates and checkpoints. Table~\ref{tab:down} lists the settings.

\begin{table}[htbp]
\caption{Downstream optimization settings.}
\label{tab:down}
\small
\begin{tabular*}{\columnwidth}{@{\extracolsep{\fill}}ll@{}}
\toprule
Setting & Value\\
\midrule
Optimizer / weight decay & AdamW~\cite{loshchilov2019adamw} / 0.01\\
Encoder learning rates & $3{\times}10^{-5}$, $10^{-4}$, $3{\times}10^{-4}$\\
Risk-head learning rate & $10^{-3}$\\
Global batch size & 128\\
Updates per LR / final run & 1,024 / 2,048\\
Warmup updates & 40\\
Schedule after warmup & Cosine~\cite{loshchilov2017sgdr} to 0.1 of initial LR\\
Gradient-norm limit & 1.0\\
Loss & Class-balanced binary cross-entropy\\
Fine-tuning runs & 3\\
\bottomrule
\end{tabular*}
\end{table}

\section{Baselines}
\label{app:objectives}
\label{app:systems}
\textbf{CoLES.} Two contiguous crops from one history form a positive pair; negatives come from other users. Unit-normalized embeddings use a Euclidean contrastive loss with margin 1 and the five closest eligible negatives. Crop lengths range from one quarter of the history to its full length, with a minimum of two events. The shared Transformer replaces the original recurrent encoder, such as an LSTM~\cite{hochreiter1997lstm}, and two crops per anchor replace lifetime-level multi-crop sampling.

\textbf{BYB.} A strictly time-separated older prefix predicts the representation of a recent event window produced by a teacher whose parameters follow an exponential moving average. The teacher updates at 0.995, and a two-layer predictor uses soft-target cross-entropy at temperature 0.1. A 60/40 event-count split with one disjoint pair per anchor replaces fixed daily windows; the teacher averages event representations instead of using daily pooling.

\textbf{CPC.} The context encoder receives only the prefix and predicts future event representations at horizons 1, 2, and 4 using horizon-specific bilinear scores and cross-user in-batch negatives. Both event and context encoders receive gradients; missing horizons are omitted. CoLES, BYB, and CPC each use 24,000 pretraining updates at batch size 128 and one pretrained checkpoint.

\textbf{TransactionGPT.} The adapted system uses two-layer metadata and feature encoders and an 18-layer causal history encoder with 640 hidden units and eight attention heads. Two virtual tokens produce a 1,280-dimensional history representation. Next-event pretraining averages categorical cross-entropy and two numerical mean-squared-error losses.

\textbf{PANTHER.} The adapted encoder uses ten layers, eight attention heads, and 64 sequence patterns. Event tokens combine a 400-dimensional behavior embedding with three 80-dimensional field embeddings. The behavior and field vocabularies contain 60,003 and 7,912 tokens, respectively. Next-event contrastive prediction uses 32 in-batch negatives and temperature 0.05.

TransactionGPT and PANTHER are pretrained from scratch on the same history pool for 24,000 updates at batch size 128 and a history cap of 300. Calendar features and user-profile modules are disabled. Adaptation uses the search and update budget in Table~\ref{tab:down}, with microbatches of 32 and a new risk head. All representation parameters are fine-tuned; pretraining-only prediction heads are frozen.

\textbf{XGBoost reference.} XGB is trained on risk indicators accumulated by Ant International's production system over several years rather than on event histories. Because its inputs differ from those of every other method, we report it as a reference (Table~\ref{tab:xgb}) rather than in the main comparison. It uses the same label budgets and weighted test metrics; the table gives its means and sample standard deviations, with the scratch and TRIDENT-Seq means from Table~\ref{tab:main} for comparison. XGB trails scratch training in all six settings; on full-label Wallet, its recall is 4.31\%, against 10.41\% for scratch and 28.33\% for TRIDENT-Seq.

\begin{table}[htbp]
\caption{XGBoost on Ant International's production risk indicators (\%).}
\label{tab:xgb}
\centering\small
\setlength{\tabcolsep}{2pt}
\begin{tabular*}{\columnwidth}{@{\extracolsep{\fill}}lrrrrrr@{}}
\toprule
&\multicolumn{3}{c}{AP}&\multicolumn{3}{c}{R@0.5\%}\\
\cmidrule(lr){2-4}\cmidrule(lr){5-7}
Setting & XGB & Scratch & Ours & XGB & Scratch & Ours\\
\midrule
Wallet 10\% & 0.0133\pmstd{0.0027} & 0.0342 & 0.8214 & 2.01\pmstd{0.37} & 3.31 & 27.75\\
Wallet 100\% & 0.1042\pmstd{0.0109} & 0.1148 & 0.8355 & 4.31\pmstd{1.16} & 10.41 & 28.33\\
Shopping 10\% & 0.2146\pmstd{0.0098} & 0.3067 & 0.4331 & 3.41\pmstd{0.94} & 4.79 & 6.61\\
Shopping 100\% & 0.3573\pmstd{0.0307} & 0.4205 & 0.5678 & 6.25\pmstd{0.41} & 7.81 & 9.61\\
SME 10\% & 0.2810\pmstd{0.0411} & 0.3476 & 0.4886 & 3.97\pmstd{0.33} & 4.67 & 9.14\\
SME 100\% & 0.5210\pmstd{0.0652} & 0.6370 & 0.8685 & 6.88\pmstd{0.25} & 8.11 & 14.34\\
\bottomrule
\end{tabular*}
\end{table}

\section{Pretraining Source Composition}
\label{app:sourcecontrols}
Table~\ref{tab:sourcecontrols} reports the controlled-source results. Each source has one pretrained checkpoint and three downstream fine-tuning runs, summarized by means and sample standard deviations. These checkpoints use the matched pools of Section~\ref{sec:sources} and are separate from the main pretraining run. The no-pretraining baseline uses the same downstream training and test sets.

\begin{table*}[t]
\caption{Cross-business transfer results (\%).}
\label{tab:sourcecontrols}
\centering\small
\setlength{\tabcolsep}{3pt}
\begin{tabular*}{\textwidth}{@{\extracolsep{\fill}}llrrrrrr@{}}
\toprule
&&\multicolumn{3}{c}{10\% labels}&\multicolumn{3}{c}{25\% labels}\\
\cmidrule(lr){3-5}\cmidrule(lr){6-8}
Target & Pretraining source & AP & AUROC & R@0.5\% & AP & AUROC & R@0.5\%\\
\midrule
Shopping & No pretraining & 0.3067\pmstd{0.0375} & 73.90\pmstd{0.66} & 4.79\pmstd{0.51} & 0.3591\pmstd{0.0597} & 74.91\pmstd{1.40} & 5.68\pmstd{1.73}\\
 & Shopping only & 0.3994\pmstd{0.0650} & 75.10\pmstd{0.56} & 5.97\pmstd{0.78} & 0.4361\pmstd{0.1080} & 74.73\pmstd{2.00} & 7.85\pmstd{2.06}\\
 & SME only & 0.3668\pmstd{0.1107} & 71.13\pmstd{6.86} & 5.96\pmstd{2.48} & 0.4663\pmstd{0.0435} & 75.57\pmstd{1.25} & 8.70\pmstd{0.76}\\
 & Shopping + SME & 0.3331\pmstd{0.1150} & 71.71\pmstd{5.84} & 6.46\pmstd{1.81} & 0.4511\pmstd{0.1033} & 74.93\pmstd{0.46} & 7.98\pmstd{1.99}\\
\midrule
SME & No pretraining & 0.3476\pmstd{0.0798} & 67.84\pmstd{13.67} & 4.67\pmstd{0.51} & 0.4103\pmstd{0.0494} & 69.52\pmstd{13.10} & 5.04\pmstd{1.25}\\
 & SME only & 0.4857\pmstd{0.0615} & 73.00\pmstd{8.24} & 6.13\pmstd{0.15} & 0.5095\pmstd{0.0883} & 58.75\pmstd{3.97} & 7.86\pmstd{0.61}\\
 & Shopping only & 0.5791\pmstd{0.0519} & 77.71\pmstd{5.43} & 8.67\pmstd{1.53} & 0.4319\pmstd{0.0543} & 65.48\pmstd{12.37} & 6.94\pmstd{0.26}\\
 & Shopping + SME & 0.4315\pmstd{0.1097} & 67.15\pmstd{17.08} & 7.15\pmstd{0.53} & 0.6379\pmstd{0.0576} & 77.18\pmstd{6.47} & 9.05\pmstd{1.73}\\
\bottomrule
\end{tabular*}
\end{table*}

\section{Data Splits and Label Budgets}
\label{app:data}
Table~\ref{tab:splits} lists the sampled split sizes and positive counts. Positive rates differ across splits because of negative sampling; validation and test metrics reweight each example by its inverse sampling probability. Label budgets subsample the training split, and Table~\ref{tab:budgetpos} gives the resulting positive counts. Wallet is evaluated at 10\% and 100\% only, because its 1\% and 5\% subsets contain 7 and 34 positives.

\begin{table}[htbp]
\caption{Sampled split sizes and positives.}
\label{tab:splits}
\small
\begin{tabular*}{\columnwidth}{@{\extracolsep{\fill}}llrrr@{}}
\toprule
Data & Split & Events & Positives & Positive rate\\
\midrule
Wallet & Train & 30,232 & 732 & 2.42\%\\
 & Validation & 46,807 & 807 & 1.72\%\\
 & Test & 50,336 & 3,336 & 6.63\%\\
\midrule
Shopping & Train & 283,040 & 183,398 & 64.80\%\\
 & Validation & 20,583 & 373 & 1.81\%\\
 & Test & 198,275 & 2,949 & 1.49\%\\
\midrule
SME & Train & 147,401 & 46,740 & 31.71\%\\
 & Validation & 20,523 & 702 & 3.42\%\\
 & Test & 203,676 & 3,381 & 1.66\%\\
\bottomrule
\end{tabular*}
\end{table}

\begin{table}[htbp]
\caption{Training positives at each label budget.}
\label{tab:budgetpos}
\small
\begin{tabular*}{\columnwidth}{@{\extracolsep{\fill}}lrrrrr@{}}
\toprule
Data & 1\% & 5\% & 10\% & 25\% & 100\%\\
\midrule
Wallet & 7 & 34 & 76 & 172 & 732\\
Shopping & 1,881 & 9,265 & 18,423 & 45,859 & 183,398\\
SME & 469 & 2,382 & 4,671 & 11,608 & 46,740\\
\bottomrule
\end{tabular*}
\end{table}

\section{Results at All Label Budgets}
\label{app:budgets}
Table~\ref{tab:budgets} reports the values plotted in Figure~\ref{fig:label} as three-run means and sample standard deviations. All methods use the same training-label subsets and the full weighted test population.

\begin{table}[htbp]
\caption{Shopping and SME results across label budgets (\%).}
\label{tab:budgets}
\centering\small
\setlength{\tabcolsep}{2pt}
\begin{tabular*}{\columnwidth}{@{\extracolsep{\fill}}lrrrr@{}}
\toprule
&\multicolumn{2}{c}{AP}&\multicolumn{2}{c}{R@0.5\%}\\
\cmidrule(lr){2-3}\cmidrule(lr){4-5}
Labels & TRIDENT-Seq & Scratch & TRIDENT-Seq & Scratch\\
\midrule
\multicolumn{5}{@{}l}{\textit{Shopping}}\\
1\% & 0.2344\pmstd{0.1087} & 0.2609\pmstd{0.0496} & 3.31\pmstd{2.18} & 4.76\pmstd{0.38}\\
5\% & 0.3546\pmstd{0.0939} & 0.3049\pmstd{0.0703} & 5.87\pmstd{1.62} & 5.11\pmstd{1.46}\\
10\% & 0.4331\pmstd{0.0735} & 0.3067\pmstd{0.0375} & 6.61\pmstd{1.16} & 4.79\pmstd{0.51}\\
25\% & 0.5246\pmstd{0.0566} & 0.3591\pmstd{0.0597} & 8.33\pmstd{1.01} & 5.68\pmstd{1.73}\\
100\% & 0.5678\pmstd{0.0487} & 0.4205\pmstd{0.0660} & 9.61\pmstd{0.73} & 7.81\pmstd{1.76}\\
\midrule
\multicolumn{5}{@{}l}{\textit{SME}}\\
1\% & 0.3684\pmstd{0.0759} & 0.2853\pmstd{0.0272} & 4.84\pmstd{0.47} & 4.50\pmstd{1.13}\\
5\% & 0.5292\pmstd{0.0097} & 0.4109\pmstd{0.0240} & 6.19\pmstd{2.04} & 5.44\pmstd{0.30}\\
10\% & 0.4886\pmstd{0.0672} & 0.3476\pmstd{0.0798} & 9.14\pmstd{1.85} & 4.67\pmstd{0.51}\\
25\% & 0.6928\pmstd{0.0965} & 0.4103\pmstd{0.0494} & 10.16\pmstd{2.05} & 5.04\pmstd{1.25}\\
100\% & 0.8685\pmstd{0.1553} & 0.6370\pmstd{0.0805} & 14.34\pmstd{0.42} & 8.11\pmstd{1.25}\\
\bottomrule
\end{tabular*}
\end{table}

\section{Ablation: AP and AUROC}
\label{app:ablation}
Table~\ref{tab:ablation-ap} reports AP and AUROC for the component variants. Full PFC has the highest AP at both label budgets. AP and top-budget recall need not move together: for the three partial variants, AP decreases from 10\% to 100\% labels while R@0.5\% increases. We observe such divergences repeatedly on Wallet, where weighted positives are rare and AP varies widely across runs (Table~\ref{tab:main}). AUROC summarizes the whole ranking, varies by less than 2.5 points across variants, and does not favor Full PFC; the components mainly change the top of the ranking, where interventions act.

\begin{table}[htbp]
\caption{Wallet ablation: AP, AUROC, and R@0.5\% (\%).}
\label{tab:ablation-ap}
\centering\small
\setlength{\tabcolsep}{2pt}
\begin{tabular*}{\columnwidth}{@{\extracolsep{\fill}}lrrrrrr@{}}
\toprule
&\multicolumn{3}{c}{10\% labels}&\multicolumn{3}{c}{100\% labels}\\
\cmidrule(lr){2-4}\cmidrule(lr){5-7}
Variant & AP & AUROC & R@0.5\% & AP & AUROC & R@0.5\%\\
\midrule
Full PFC & \textbf{0.8373} & 91.46 & \textbf{24.62} & \textbf{1.2725} & 91.63 & \textbf{28.44}\\
PFC only & 0.5913 & 90.33 & 14.37 & 0.3703 & 91.05 & 16.07\\
w/o activity & 0.6508 & 91.41 & 15.10 & 0.4399 & 92.12 & 16.38\\
w/o dispersion & 0.7921 & \textbf{92.05} & 16.14 & 0.4153 & \textbf{93.35} & 17.21\\
\bottomrule
\end{tabular*}
\end{table}

\section{Longer Adaptation}
\label{app:extended}
On full-label Wallet, each learning-rate trial receives 2,048 updates and the final run receives 8,192 updates, giving 14,336 selection-and-training updates per method and run. Table~\ref{tab:extended} reports means and sample standard deviations over three runs, scaled so that the scratch mean under this schedule is 100 for each metric.

\begin{table}[htbp]
\caption{Longer adaptation on full-label Wallet, relative to scratch.}
\label{tab:extended}
\centering\small
\begin{tabular*}{\columnwidth}{@{\extracolsep{\fill}}lrrr@{}}
\toprule
Model & AP $\uparrow$ & AUROC $\uparrow$ & R@0.5\% $\uparrow$\\
\midrule
Scratch & 100.0\pmstd{156.1} & 100.0\pmstd{25.4} & 100.0\pmstd{36.1}\\
TRIDENT-Seq & 178.8\pmstd{94.8} & 115.4\pmstd{18.2} & 252.7\pmstd{48.4}\\
\bottomrule
\end{tabular*}
\end{table}

\section{Case Inspection Details}
\label{app:case}
The region in Figure~\ref{fig:proj} is a post-hoc 55\%-peak class-density contour on a fixed 140$\times$140 Gaussian KDE grid over t-SNE. It includes 374 histories; the 2,312-history comparison set excludes the surrounding island. Feedback statistics use 361 and 916 annotated event pairs, respectively. Merchant links and outcomes come from event inspection and are not encoder inputs.

\section{Reporting Delays and SME CB Estimates}
\label{app:onlinecheck}
Table~\ref{tab:delays} reports business-reporting delays for April--May 2026 transaction cohorts, using labels observed in the October 26 extract. Percentages count valid observed labels; future unreported labels are excluded. Fraud union combines bank-reported fraud and fraud chargebacks.

\begin{table}[htbp]
\caption{Observed label-reporting delays (\%); D95/D99 in days.}
\label{tab:delays}
\centering\small\setlength{\tabcolsep}{2pt}
\begin{tabular*}{\columnwidth}{@{\extracolsep{\fill}}llrrrrrr@{}}
\toprule
Cohort & Label & 30d & 60d & 90d & 120d & D95 & D99\\
\midrule
April & Union & 67.16 & 85.29 & 93.54 & 97.90 & 99 & 148\\
 & CB & 65.41 & 85.31 & 94.34 & 99.37 & 93 & 119\\
May & Union & 64.09 & 86.06 & 94.28 & 98.70 & 92 & 124\\
 & CB & 63.02 & 85.34 & 94.51 & 99.43 & 92 & 117\\
Apr--May & Union & 65.53 & 85.70 & 93.93 & 98.32 & 95 & 135\\
 & CB & 64.14 & 85.32 & 94.43 & 99.40 & 92 & 119\\
\bottomrule
\end{tabular*}
\end{table}

Table~\ref{tab:datecheck} uses 66 common SME transaction dates, with amounts recorded by October 25. Pooled rates divide total CB by total successful amount. Date-weighted rates average daily CB rates using combined daily successful amounts across arms as common weights. Approximate 95\% CIs use 50,000 paired resamples of 10 calendar-week blocks for the observed snapshot.

\begin{table}[!htbp]
\caption{SME transaction-date CB reduction versus XGB (\%); approximate 95\% CIs in brackets.}
\label{tab:datecheck}
\centering\small
\begin{tabular*}{\columnwidth}{@{\extracolsep{\fill}}lrr@{}}
\toprule
Method & Pooled & Date-weighted\\
\midrule
SAINT-style & \makecell[r]{12.66\\{[2.58, 22.75]}} & \makecell[r]{12.49\\{[1.43, 23.16]}}\\
TRIDENT-Seq & \makecell[r]{28.28\\{[10.69, 42.38]}} & \makecell[r]{26.09\\{[9.24, 41.34]}}\\
\bottomrule
\end{tabular*}
\end{table}

\end{document}